\documentclass[journal=ancac3,manuscript=article]{achemso}

\usepackage{amsmath}
\usepackage{graphicx}
\usepackage[version=4]{mhchem}
\usepackage{booktabs}
\usepackage{siunitx}
\DeclareSIUnit{\Molar}{M}
\usepackage{tikz}
\usepackage{pgfplots}
\pgfplotsset{compat=1.18} 
\usetikzlibrary{positioning, fit, patterns}
\usepackage{xcolor}
\usepackage{multirow}

\usepackage{layouts}

\title{Deep Learning-Driven Peptide Classification in Biological Nanopores}%

\author{Julian Ho{\ss}bach}%
\altaffiliation{These authors contributed equally to this work.}
\affiliation{Institute for Computational Physics, University of Stuttgart, 70569 Stuttgart, Germany}
\author{Samuel Tovey}%
\altaffiliation{These authors contributed equally to this work.}
\affiliation{Institute for Computational Physics, University of Stuttgart, 70569 Stuttgart, Germany}
\author{Sandro Kuppel}%
\affiliation{Institute for Computational Physics, University of Stuttgart, 70569 Stuttgart, Germany}
\author{Tobias Ensslen}%
\author{Jan C. Behrends}%
\affiliation{Laboratory for Membrane Physiology and Technology, Department of Physiology, Faculty of Medicine, University of Freiburg, 79104 Freiburg, Germany.}
\author{Christian Holm}%
\email{holm@icp.uni-stuttgart.de}
\affiliation{Institute for Computational Physics, University of Stuttgart, 70569 Stuttgart, Germany}


\begin{document}

\onehalfspacing

\begin{abstract}
Nanopore-based single-molecule sensing is a promising route to fast, low-cost disease diagnosis and protein sequencing: as an analyte such as a peptide or protein traverses a nanoscale pore, it modulates the ionic current, producing a resistive pulse whose signature is determined by the analyte's structure and its interactions with the pore.
Translating these signatures into reliable molecular identities, however, is an open problem well suited for machine learning, as the signals are noisy, suffer from variations due to experimental conditions, and are difficult to featurize, which has so far limited classification accuracy.
Here we translate the peptide identification problem into an image-classification task by transforming each resistive pulse into a scaleogram via the continuous wavelet transform, a representation that jointly encodes amplitude, frequency, and time in a form well suited for deep convolutional models.
On a dataset of 42 peptides, recorded as six separate peptide ladders, this approach reaches a macro-averaged classification accuracy of $82\,\%$ on held-out events, an improvement of $8.6$ percentage points over the descriptor-based approach previously reported for the same dataset.
We further show that the trained models tolerate substantial compression, retaining their accuracy with half of their weights set to zero and under 8-bit quantization, a prerequisite for deploying trained classifiers on embedded sensing hardware.
Our results demonstrate how physically motivated signal representations can make complex single-molecule data tractable for modern learning algorithms, a step on the path towards point-of-care peptide and protein diagnostics.
\end{abstract}


\section*{Background}\label{sec:background}
Recent developments in cancer research have indicated that sequences of amino acids within a biological sample can serve as early indicators for certain cancers, potentially years before initial diagnosis~\citep{papier24a}.
This discovery opens a promising avenue for early cancer detection by the recognition of these amino acid segments, typically found in proteins or peptides within biological samples.
Indicators can manifest as elevated levels of common proteins or the presence of post-translational modifications (PTMs) \textemdash changes to a protein or peptide after biosynthesis.
\citet{barker23a} highlight current research stating that as of 2022, 3224 peptides along with 1742 proteins have been validated by the Clinical Proteomic Tumor Analysis Consortium (CPTAC)~\citep{whiteaker14a} for use in the development of reproducible assays for cancer biomarkers~\cite{son19a, rudnik16a}.
For this reason, fast, low-cost, accurate methods for identifying large numbers of proteins, peptides, and PTMs with single-molecule precision are desired.
Traditional methods for protein identification include mass spectrometry~\citep{cox11a} on samples containing thousands of proteins and peptides or on individual proteins separated prior analysis by electrophoresis~\citep{nowakowski14a} or chromatography.
In either case, the process is expensive, time-consuming and importantly, requires processing in a laboratory separated from clinicians, thus impeding real time, on-premises diagnosis.
Nanopore-based classification devices hold significant potential to overcome these limitations~\citep{wang21a, ying22a}.
Nanopores are single nanometer scale openings in an otherwise insulating membrane, embedded in a saline solution.
A constant ion current flowing through the pore can be measured as a consequence of an externally applied electric field across the membrane.
Analyte molecules entering such pore alter this ionic current characteristically, thereby providing information regarding their identity.
These alterations be transient current reductions, referred to as resistive pulses.
Nanopores can be solid-state~\citep{dekker07a}, such as holes in graphene sheets, or biological~\citep{cao16a}, built from proteins, e.g. Aerolysin, a pore-forming toxin produced by \emph{Aeromonas hydrophila}, reconstituted in lipid membranes.
Due to the simplicity of this apparatus, nanopore devices are cheap, small and can be operated with very little pre-processing, making them a promising candidate for clinical applications.

Nanopore technology first emerged as a means for sequencing DNA, whereby a pore is used as a single molecule sensor, and a motor enzyme is placed over the pore to control translocation speed of an extended DNA strand in a single base ratchet motion~\citep{schneider12a}.
The bases present inside the channel's constriction block the pore current to a characteristic current that can then be used to identify them~\citep{winston12a,boza17a,noakes19a,carral21a,neumann22}.
The directed progression of the DNA molecule therefore allows sequencing of the entire strand.
As this approach proved useful for DNA sequencing, a direct translation of the approach to the more complex case of proteins or peptides faces several limitations, the first being the absence of a directed, controlled ratcheting mechanism for (poly)aminoacids, although current research attempts to overcome this limitation~\citep{brinkerhoff21a}.

This work relies on a different mechanism for detection of single molecules by nanopores, referred to as single molecule trapping.
In contrast to sequentially feeding an elongated polymer through the pore~\citep{kasianowicz96a}, the entire analyte molecule is entrapped in a local minimum of potential energy inside the pore for a period of time before leaving it stochastically~\citep{robertson07a}.
As a consequence, the resistive pulse entails information of the entire molecule instead of solely a fraction.
The second challenge arises from the complexity of the proteins and peptides.
Unlike DNA, which consist of four different base pairs, each being chemically relatively similar, peptides can be built from 20 different kinds of amino acids, spanning a wide range of different physicochemical properties.
These factors result in non-trivial blockade currents, impacted by the spatial orientation of the peptide inside the pore, interactions of it with the pore walls, as well as with charge carriers in the solution.
Instead of measuring a multitude of complex strand-fractions, typically hampering successful sequencing~\citep{wang23b,he21b,ritmejeris24a}, trapping of entire molecules reduces the signal complexity significantly, thereby potentially facilitating peptide discrimination, as shown by previous studies.

\citet{ouldali20a} and~\citet{bakshloo22a} have successfully demonstrated the mean blockade current being primarily related to the volume of the peptide in the pore, leading to discrimination of peptides with single amino acid differences, or in an outstanding case~\citep{ensslen22a} even positional isomers of post translationally modified peptides reveling the same mass and overall volume.
Using machine learning–based methods, single–amino acid resolution has been achieved by extracting informative features from blockade current signals and applying classifiers such as support vector machines and neural networks~\citep{wang24a, cao24a, zhang24a, wang25a}.
These approaches have also been successfully applied to multi-pass protein read experiments~\citep{motone24a}.
In the context of large-scale peptide classification, the authors previously reported the classification of up to 42 peptides with a macro-averaged accuracy of $73.6\,\%$, relying on statistical moments and catch22 features of blockade current distributions~\citep{hossbach25a}.
While peptides of differing lengths were discriminated reliably by the used feature sets as well as with classical histogram-based analysis, peptides of similar lengths posed difficulties to classification to both methods.
To further improve the classification accuracy, this work introduces a new approach to peptide classification in nanopore experiments and evaluates its performance on the same dataset by leveraging the image classification capabilities of deep neural networks.
Specifically, we convert resistive pulses of single blockade events into so-called \emph{scaleogram} images via wavelet transformation~\citep{hossain99a}, capturing time, frequency, and amplitude information.
Further, deep vision models, including a ResNet18~\citep{he15a}, ResNeXt101~\citep{xie17a}, and a Vision Transformer (ViT)~\citep{dosovitskiy21a} are then trained on the classification task, increasing the overall classification accuracy up to $82.2\,\%$ on the 42 peptides.
The results outlined here improve on the highest accuracy previously reported for this dataset and show that image-based deep learning is a viable route to automated peptide classification in nanopore recordings, a prerequisite for the eventual clinical application of this technology.

\section*{Methods}\label{sec:methods}
\subsection*{Peptide ladder}
This work applies classification models to 42 peptides in the form of peptide ladders~\citep{behrends22a}.
These 42 peptides arise from 6 model decapeptides, each equipped with a C-terminal tri-arginine handle to maintain orientation of entry into the pore.
Notably, all 6 decapeptides share the same amino acid composition in their N-terminal heptapeptide section (\ce{AKR2S2Y}) but differ in sequence.
N-terminal truncation of each decapeptide \ce{X_{7-1}R3} down to \ce{R3}, leading to so called peptide ladders, resulted in a total number of 42 peptides.
Figure~\ref{fig:peptide-ladders} reveals the structure of the six peptide ladders used in the study to derive the ladders.
The far left side shows the N-terminus indicated by H, and the far right shows the C-terminus denoted as \ce{OH}.
Letters S, R, A, K, and Y correspond to the amino acids Serine, Arginine, Alanine, Lysine, and Tyrosine.

\begin{figure*}[ht]
	\definecolor{color1}{HTML}{E2C2FF}
\definecolor{color2}{HTML}{CBD4C2}
\definecolor{color3}{HTML}{FC440F}
\definecolor{color4}{HTML}{247BA0}
\definecolor{color5}{HTML}{C3B299}

\begin{tikzpicture}[
    ladder/.style={rectangle, draw, minimum size=1cm, font=\small},
    acidS/.style={draw, rounded corners, minimum width=1cm, align=center, fill=color1},
    acidR/.style={draw, rounded corners, minimum width=1cm, align=center, fill=color2},
    acidA/.style={draw, rounded corners, minimum width=1cm, align=center, fill=color3},
    acidY/.style={draw, rounded corners, minimum width=1cm, align=center, fill=color4},
    acidK/.style={draw, rounded corners, minimum width=1cm, align=center, fill=color5},
    terminus/.style={font=\bfseries, anchor=west},
    faint/.style={draw, rounded corners, minimum width=1cm, align=center, fill=gray!20},
    checkered/.style={draw, rounded corners, minimum width=1cm, align=center, pattern=north east lines, pattern color=gray!60}
]

\node[terminus] (L1H) at (0,0) {H};
\node[acidS, right=0.3cm of L1H] (L1S1) {S};
\node[acidR, right=0.3cm of L1S1] (L1R) {R};
\node[acidA, right=0.3cm of L1R] (L1A) {A};
\node[acidS, right=0.3cm of L1A] (L1S2) {S};
\node[acidK, right=0.3cm of L1S2] (L1K) {K};
\node[acidY, right=0.3cm of L1K] (L1Y) {Y};
\node[acidR, right=0.3cm of L1Y] (L1R2) {R};
\node[acidR, right=0.3cm of L1R2] (L1R3) {R};
\node[acidR, right=0.3cm of L1R3] (L1R4) {R};
\node[acidR, right=0.3cm of L1R4] (L1R5) {R};
\node[checkered, fit={(L1R3) (L1R4) (L1R5)}, inner sep=0.2cm] (box) {};
\node[terminus, right=0.3cm of L1R5] (L1OH) {OH};

\node[terminus, below=0.6cm of L1H] (L2H) {H};
\node[acidK, right=0.3cm of L2H] (L2K1) {K};
\node[acidS, right=0.3cm of L2K1] (L2S1) {S};
\node[acidR, right=0.3cm of L2S1] (L2R1) {R};
\node[acidA, right=0.3cm of L2R1] (L2A1) {A};
\node[acidS, right=0.3cm of L2A1] (L2S2) {S};
\node[acidR, right=0.3cm of L2S2] (L2R2) {R};
\node[acidY, right=0.3cm of L2R2] (L2Y1) {Y};
\node[acidR, right=0.3cm of L2Y1] (L2R3) {R};
\node[acidR, right=0.3cm of L2R3] (L2R4) {R};
\node[acidR, right=0.3cm of L2R4] (L2R5) {R};
\node[checkered, fit={(L2R3) (L2R4) (L2R5)}, inner sep=0.2cm] (box) {};
\node[terminus, right=0.3cm of L2R5] (L2OH) {OH};

\node[terminus, below=0.6cm of L2H] (L3H) {H};
\node[acidR, right=0.3cm of L3H] (L3S1) {R};
\node[acidY, right=0.3cm of L3S1] (L3R) {Y};
\node[acidS, right=0.3cm of L3R] (L3A) {S};
\node[acidR, right=0.3cm of L3A] (L3S2) {R};
\node[acidA, right=0.3cm of L3S2] (L3K) {A};
\node[acidS, right=0.3cm of L3K] (L3Y) {S};
\node[acidK, right=0.3cm of L3Y] (L3R2) {K};
\node[acidR, right=0.3cm of L3R2] (L3R3) {R};
\node[acidR, right=0.3cm of L3R3] (L3R4) {R};
\node[acidR, right=0.3cm of L3R4] (L3R5) {R};
\node[checkered, fit={(L3R3) (L3R4) (L3R5)}, inner sep=0.2cm] (box) {};
\node[terminus, right=0.3cm of L3R5] (L3OH) {OH};

\node[terminus, below=0.6cm of L3H] (L4H) {H};
\node[acidK, right=0.3cm of L4H] (L4S1) {K};
\node[acidS, right=0.3cm of L4S1] (L4R) {S};
\node[acidR, right=0.3cm of L4R] (L4A) {R};
\node[acidY, right=0.3cm of L4A] (L4S2) {Y};
\node[acidA, right=0.3cm of L4S2] (L4K) {A};
\node[acidR, right=0.3cm of L4K] (L4Y) {R};
\node[acidS, right=0.3cm of L4Y] (L4R2) {S};
\node[acidR, right=0.3cm of L4R2] (L4R3) {R};
\node[acidR, right=0.3cm of L4R3] (L4R4) {R};
\node[acidR, right=0.3cm of L4R4] (L4R5) {R};
\node[checkered, fit={(L4R3) (L4R4) (L4R5)}, inner sep=0.2cm] (box) {};
\node[terminus, right=0.3cm of L4R5] (L4OH) {OH};

\node[terminus, below=0.6cm of L4H] (L5H) {H};
\node[acidK, right=0.3cm of L5H] (L5S1) {K};
\node[acidR, right=0.3cm of L5S1] (L5R) {R};
\node[acidS, right=0.3cm of L5R] (L5A) {S};
\node[acidS, right=0.3cm of L5A] (L5S2) {S};
\node[acidR, right=0.3cm of L5S2] (L5K) {R};
\node[acidA, right=0.3cm of L5K] (L5Y) {A};
\node[acidY, right=0.3cm of L5Y] (L5R2) {Y};
\node[acidR, right=0.3cm of L5R2] (L5R3) {R};
\node[acidR, right=0.3cm of L5R3] (L5R4) {R};
\node[acidR, right=0.3cm of L5R4] (L5R5) {R};
\node[checkered, fit={(L5R3) (L5R4) (L5R5)}, inner sep=0.2cm] (box) {};
\node[terminus, right=0.3cm of L5R5] (L5OH) {OH};

\node[terminus, below=0.6cm of L5H] (L6H) {H};
\node[acidS, right=0.3cm of L6H] (L6S1) {S};
\node[acidK, right=0.3cm of L6S1] (L6R) {K};
\node[acidR, right=0.3cm of L6R] (L6A) {R};
\node[acidY, right=0.3cm of L6A] (L6S2) {Y};
\node[acidS, right=0.3cm of L6S2] (L6K) {S};
\node[acidR, right=0.3cm of L6K] (L6Y) {R};
\node[acidA, right=0.3cm of L6Y] (L6R2) {A};
\node[acidR, right=0.3cm of L6R2] (L6R3) {R};
\node[acidR, right=0.3cm of L6R3] (L6R4) {R};
\node[acidR, right=0.3cm of L6R4] (L6R5) {R};
\node[checkered, fit={(L6R3) (L6R4) (L6R5)}, inner sep=0.2cm] (box) {};
\node[terminus, right=0.3cm of L6R5] (L6OH) {OH};

\end{tikzpicture}
	\caption{
		Visualization of the peptide ladders used in the investigation.
		For each ladder, amino acids can be cleaved individually starting from the N-terminus up to the three Arginine (R) bases on the C-terminus.
		Thus, for each complete ladder, 6 smaller sub-ladders can be constructed, providing a total of 7 samples.
		Given the 6 available ladders, this results in 42 total peptides.
	}
	\label{fig:peptide-ladders}
\end{figure*}

\subsection*{Nanopore recordings and data hierarchy}\label{subsec:experiment-methods}
All current traces analyzed in this work were recorded with a aerolysin (AeL) nanopore following the protocols reported previously~\citep{ensslen22a, hossbach25a}.
Measurements were performed on a modified Orbit 16 platform (Nanion Technologies, Munich, Germany) in combination with MECA16 microelectrode cavity arrays (Ionera Technologies, Freiburg, Germany).
One of the 16 coplanar gold lines contacting the \SI{50}{\micro\meter} cavities was connected to the headstage of an Axopatch 200B patch-clamp amplifier (Molecular Devices, Sunnyvale, CA) through a \SI{1}{\centi\meter} unshielded silver wire.
The amplifier was operated in capacitive feedback mode at a gain of \SI{50}{\milli\volt\per\pico\ampere} with its internal low-pass filter set to a \SI{100}{\kilo\hertz} cutoff (\SI{-3}{\decibel}).
The output was passed through an external eight-pole Bessel low-pass filter (npi electronic, Tamm, Germany, custom LHBF-48X-8HL) with a \SI{50}{\kilo\hertz} cutoff and digitized at \SI{1}{\mega\hertz} with a PCI-6251 16-bit ADC interface (National Instruments, Austin, TX) driven by the GePulse software (Michael Pusch, University of Genoa, Italy).
Electrodes were immersed in \SI{150}{\micro\liter} of \SI{4}{\Molar} KCl and lipid bilayers were formed over the cavity orifices by bubble painting.
Stable single AeL pores were obtained in most membranes within ten minutes of adding activated pro-aerolysin.
The peptides were obtained by solid-phase custom synthesis (Intavis Peptide Services GmbH, Tübingen, Germany) as a lyophilized powder, dissolved in \SI{5}{\milli\Molar} HEPES at pH $7.5$, and used at a final concentration of \SI{1}{\micro\Molar}.
Recordings were exported in ABF format for the analysis described below.
Each of the six ladders, L1--L6, was measured in a separate experiment using a single pore in a single MECA16 chip on a single day, with the multi-peptide mixture of that ladder established \textit{in situ} by sequential addition from individual stock solutions, so that pore, chip, recording session and preparation day coincide with ladder identity.

\subsection*{Event detection and labeling}\label{subsec:event-methods}
During nanopore experiments, a time series of current measurements is captured, within which many events will occur, as shown in Figure~\ref{fig:data-processing} a).
Before these events can be classified, they are isolated from the long current readout into single events and stored as single events.
As a pre-processing step, a wavelet threshold filter using the \verb|bior1.5| wavelet and a hard threshold of $0.5$ was applied to the raw data.
A statistical threshold detection approach~\citep{piguet21a, cao24a} is then performed, wherein deviations from the Gaussian-like distribution of the open-pore current are used to detect the onset of an event.
During event detection, the signal is processed in series, computing a windowed average over the pore current to allow for changes in the open-pore values.
If the current falls below three standard deviations, $\sigma$, from the open-pore value, it is considered to be the start of an event.
If an event is found to have a mean current below a second threshold of $4\sigma$, it is ignored as an outlier, in line with the first method introduced in~\citet{piguet21a}.
Further reduction of spurious events was achieved by only allowing events with a dwell time greater than \SI{80}{\micro\second}, attributing for the instrumentation's temporal resolution.
Finally, all detected events were normalized by the mean of the surrounding open pore current.

Histograms of the current values from single ladder experiments can then be analyzed and labeled.
Labeling is performed by first fitting the peaks of the histograms with Voigt profiles~\citep{garcia06a}.
Peptide-specific events were then identified if they fell within each profile's full-width-half-maximum (FWHM).
The FWHM intervals obtained from the fits are disjoint, so that every event is assigned at most one label.
In practice, each ladder, L1-6, was studied individually.
Labeling is therefore performed using the normalized blockade current histogram for all measurements of a single ladder, so that by construction of the labeling scheme no event can be assigned to a peptide from a different ladder.
Confusion between peptides of equal length from different ladders, i.e., those with the most significant histogram overlap, is thus excluded by the experimental design rather than resolved by the labeling procedure, which also means that the procedure cannot detect errors of this type.
The residual labeling error consequently stems only from the overlap of the Voigt profiles of sub-ladders within a single ladder, as can be seen in Figure~\ref{fig:data-processing} d).
We quantified it by computing, for each label, the fraction of the fitted Voigt area within its FWHM window that is contributed by its own peak rather than by a neighboring peak of the same ladder, yielding a mean label confidence of $97.4\,\%$.
The derivation of this measure and the per-label confidences are given in the Supplementary Information.

\subsection*{Scaleogram image construction}\label{subsec:image-generation}
A scaleogram is an image formed by performing a wavelet transformation on raw time-series data~\citep{sejdic08a} by
\begin{equation}
	W(a, b) = \frac{1}{\sqrt{|a|}} \int_{-\infty}^{\infty} x(t) \, \psi^*\left(\frac{t-b}{a}\right) dt
	\label{eqn:mother}
\end{equation}
with $\psi(x)$ the mother wavelet, as shown in Figure~\ref{fig:data-processing} c) with the time parameter (b) on the $x$-axis and frequency, $a$ on the $y$-axis.
Such an operation produces an image where each pixel corresponds to a single $(a, b)$ coordinate and color corresponding to the solution of Equation~\ref{eqn:mother}.
Thus, after a wavelet transform, all of an event's time, frequency, and amplitude information is converted into image data, for which machine-learning approaches are particularly well suited.
In the investigations, the \textit{ssqueezepy} Python library~\citep{muradeli20a} was used to perform the wavelet transforms.
Models were trained using the mother wavelets \verb|cmhat|, \verb|hhhat|, \verb|gmw|, \verb|morlet| and \verb|bump|.
The Vision Transformer was trained only on the \verb|hhhat| wavelet.
Our results show no clear advantage over any single wavelet during training, as all mother wavelets yielded comparable training performance.
We expect more training data and further optimization of the wavelet to enhance classification performance, and optimizing this wavelet is subject to future research.
Images were then resized to a width and height of 224 pixels through a bilinear interpolation.

\subsection*{Classifier models}\label{subsec:classifiers}
This study explores the use of images for peptide classification and requires neural network architectures capable of processing these images.
The full dataset consists of approximately $374,000$ images containing 42 classes to be classified.
This data was first split into a test partition containing $15\,\%$ of all events, performed at the level of individual events, drawn at random while preserving the class imbalance.
The test partition was generated with a fixed random seed and is identical across all architectures, wavelets and initializations reported below.
The remainder was divided into $80\,\%$ training and $20\,\%$ validation data using different seeds for each model run, corresponding to approximately $56,000$ test, $254,000$ training and $64,000$ validation images.
In this study, the ResNet18~\citep{he15a}, ResNeXt101~\citep{xie17a}, and a Vision Transformer~\citep{dosovitskiy21a}, the parametrization of which is presented in Table~\ref{tab:net-arch}, were chosen due to their wide acceptance in the community.
\begin{table}[ht]
	\caption{Network architectures used in the study.}
	\label{tab:net-arch}
	\begin{tabular}{@{}lcc@{}}
		\toprule
		\multicolumn{1}{c}{Model Name} & Layer Type    & Number of Parameters \\ \midrule
		ResNet18                       & Convolutional & 11,194,922           \\
		ResNeXt101-64x4d               & Convolutional & 81,489,194           \\
		Vision Transformer             & Transformer   & 9,634,218            \\ \bottomrule
	\end{tabular}
\end{table}
All models were implemented within the PyTorch framework~\citep{paszke19a} and trained on two NVIDIA L4 GPUs.
PyTorch Lightning was used for the multi-device training routine~\citep{falcon19a}.
For all models, input images were normalized based on the mean and standard deviation of all pixel values in the training dataset to avoid potential data leakage.
During training, the inputs were subject to image augmentations: random erasure~\citep{zhong17a} with probability $0.1$, horizontal and vertical flips each with probability $0.2$, and random resize-cropping to $224\times224$ pixels with a scale range of $0.7$ to $1.0$.
These augmentations were applied to the training split only, validation and test images were normalized but otherwise left unaltered.
Image augmentations of this kind have been shown to improve model generalization and have the effect of artificially expanding the dataset~\citep{nagaraju22a}.
Furthermore, all models were trained to minimize a categorical cross-entropy loss function as implemented in PyTorch.
Both the ResNet18 and ResNeXt101 architectures were initialized using pre-trained weights from models trained on the ImageNet dataset~\citep{deng09a} as this has been shown to improve classification performance~\citep{hendrycks19a}.
The ResNet18 and ResNeXt101 models were trained using stochastic gradient descent with a momentum of $0.9$~\citep{sutskever13a} and a weight decay of $10^{-4}$, and a multi-step learning rate scheduler was deployed.
Due to the size of the dataset and the models, the ResNeXt101 and ResNet18 were trained using 100 mini-batches of 50 images, resulting in an effective batch size of $5000$ images.
Finally, Stochastic Weight Averaging~\citep{izmailov19a} was employed as implemented in the PyTorch Lightning package, as it has been shown to improve performance after initial training~\citep{athiwaratkun19a}.
The models were allowed to train for at most $1500$ epochs, with training stopped manually when no further improvement was expected.
For each run, the checkpoint achieving the highest validation accuracy was retained, and all results reported here were computed with that checkpoint.

The Vision Transformer utilized in this work consists of twelve transformer blocks, each with four attention heads.
To construct the input to the transformer, 14x14 pixel blocks were flattened and passed through a single dense network layer, projecting them into a 256-dimensional representation.
An additional learnable positional embedding was applied to correlate the patches in the transformer.

\begin{figure*}[ht]
	\centering
	\includegraphics[width=0.7\linewidth]{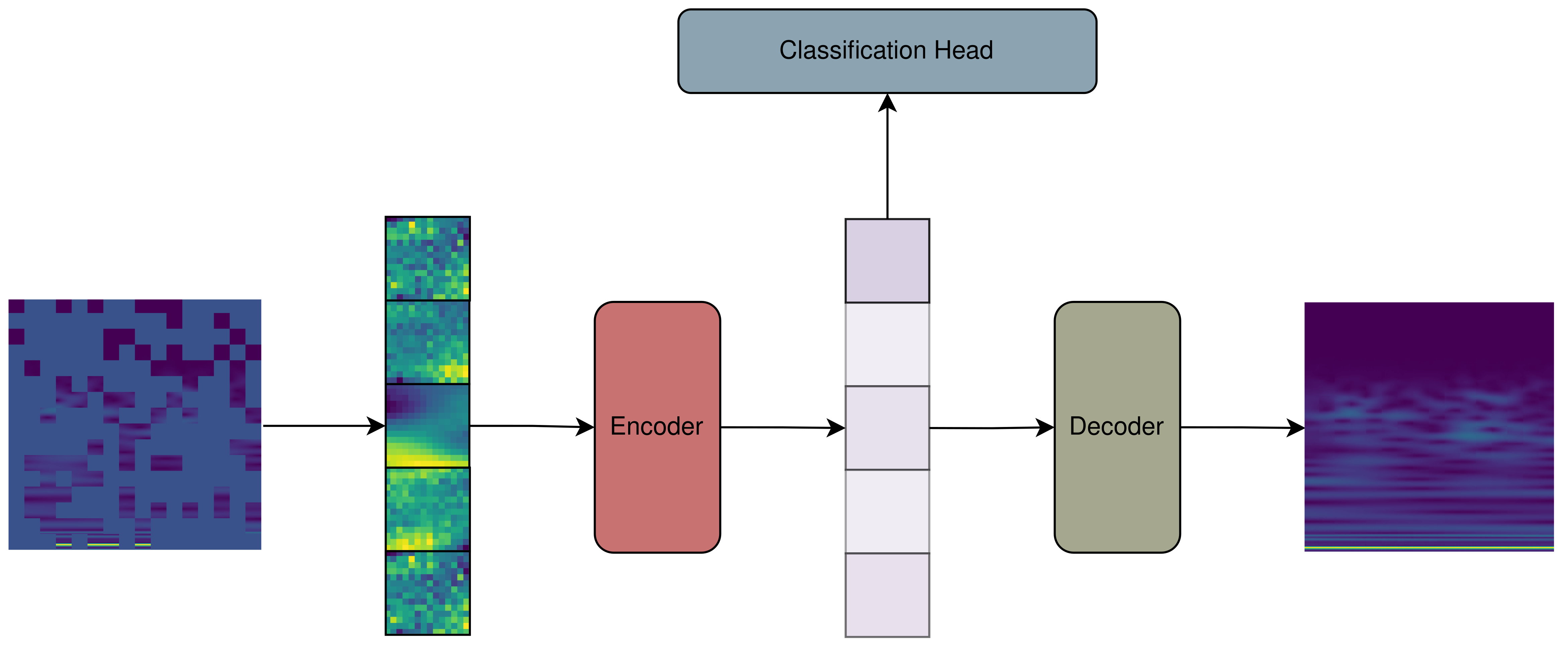}
	\caption{The encoder-decoder architecture used in pre-training the Vision Transformer. Initially, a wavelet image is divided into input patches. During pre-training, $70\,\%$ of the patches are masked, while the remaining ones are passed into the encoder. The encoder generates a latent space representation, which is then concatenated with the masked patches and passed through the decoder, aiming to reconstruct the original image. After pre-training, the classification head uses the pre-trained latent space representation for the classification task.}
	\label{fig:end-dec-architecture}
\end{figure*}

Training the Vision Transformer was performed in two steps, in line with the work from which the architecture is taken~\citep{he21a}.
The first step involves unlabeled pre-training, as outlined in Figure~\ref{fig:end-dec-architecture}.
During this phase, $70\,\%$ of each image is masked using a learnable mask token, with the remaining visible patches passed into the transformer encoder.
The output of this pass is then concatenated with the mask tokens in the original order and passed through a decoder block of the same architecture as the transformer encoder.
This decoder block produces a reconstructed image that is then compared to the original, unmasked image.
The model was trained to reconstruct the full image accurately from the sparse chunks it receives correctly.
Training took place with an effective batch size of 1600 images split into eight minibatches of 200.
The learning rate was set first by a linear warm-up to an Adam optimizer, followed by a cosine decay~\citep{loshchilov17a} with a max value of $10^{-4}$ throughout the rest of the training.
This pre-training has the effect of tuning the latent space representation of the encoder transformer, which will later be used for classification, such that it can completely reconstruct its input, an approach shown on many occasions to improve classification performance~\citep{erhan10a, devlin19a}.
Further, pre-training does not require labeled data.
Pre-training, fine-tuning and evaluation used the same partition of the data, so that no event reserved for validation or testing was seen by the encoder at any stage of training, even in unlabeled form.

Once pre-training was completed, the classification head, a one-layer dense network with GELU activation function~\citep{hendrycks23a} of architecture $\left(\mathcal{D}^{128}\cdot\text{GELU}\cdot\mathcal{D}^{42}\right)$, was trained on the classification task, taking the latent space generated by the encoder as input.
As fine-tuning the encoder's latent space representation during the classification task further improves performance, both the classification head and the transformer encoder were updated during training.
Only the initial projection of the input patches into the latent space dimension was frozen after pre-training.
During this stage of training, a batch size of 800 images split into mini-batches of 200 was used over 360 epochs.
The learning rate followed the same procedure outlined in the pre-training.

\subsection*{Importance analysis}\label{subsec:shap}
The importance analysis of the scaleograms was implemented using the Captum framework~\citep{kokhlikyan20a-pre}.
Attributions to the input images were calculated using the DeepLiftSHAP algorithm.
As a comparison baseline, wavelet scaleograms of the open pore current were computed, subject to the same normalization and resizing operations as the test dataset.
The open pore current was chosen as the reference because it is the signal recorded in the absence of a peptide, so that the resulting attributions quantify which regions of a scaleogram distinguish an event from the stochastic open pore signal.
The lengths of the open pore current slices were chosen by uniformly sampling from the distribution of dwell times of all events in the dataset.
This was done to construct a comparative baseline that is not biased towards specific dwell times.
Attributions were computed for random samples from the test dataset.

\subsection*{Weight pruning and quantization}\label{subsec:model-transfer-methods}
This work implemented the global unstructured pruning procedure in the PyTorch framework~\citep{paszke19a}.
In this approach to pruning, the parameters of all dense and convolutional layers were subject to L1 unstructured pruning.
To do so, the magnitude of each weight was measured, and, depending on the pruning fraction, $f$, the most negligible $f\,\%$ of weights were set to zero.
Additional work should explore the role of more advanced, possibly structured pruning and quantization approaches.
However, it's crucial to prioritize the issue of increased training data, as it directly impacts the performance and scalability of models.

\section*{Results}\label{sec:results}
\subsection*{Data collection and processing}

\begin{figure*}[ht]
	\centering
	\includegraphics[width=\linewidth]{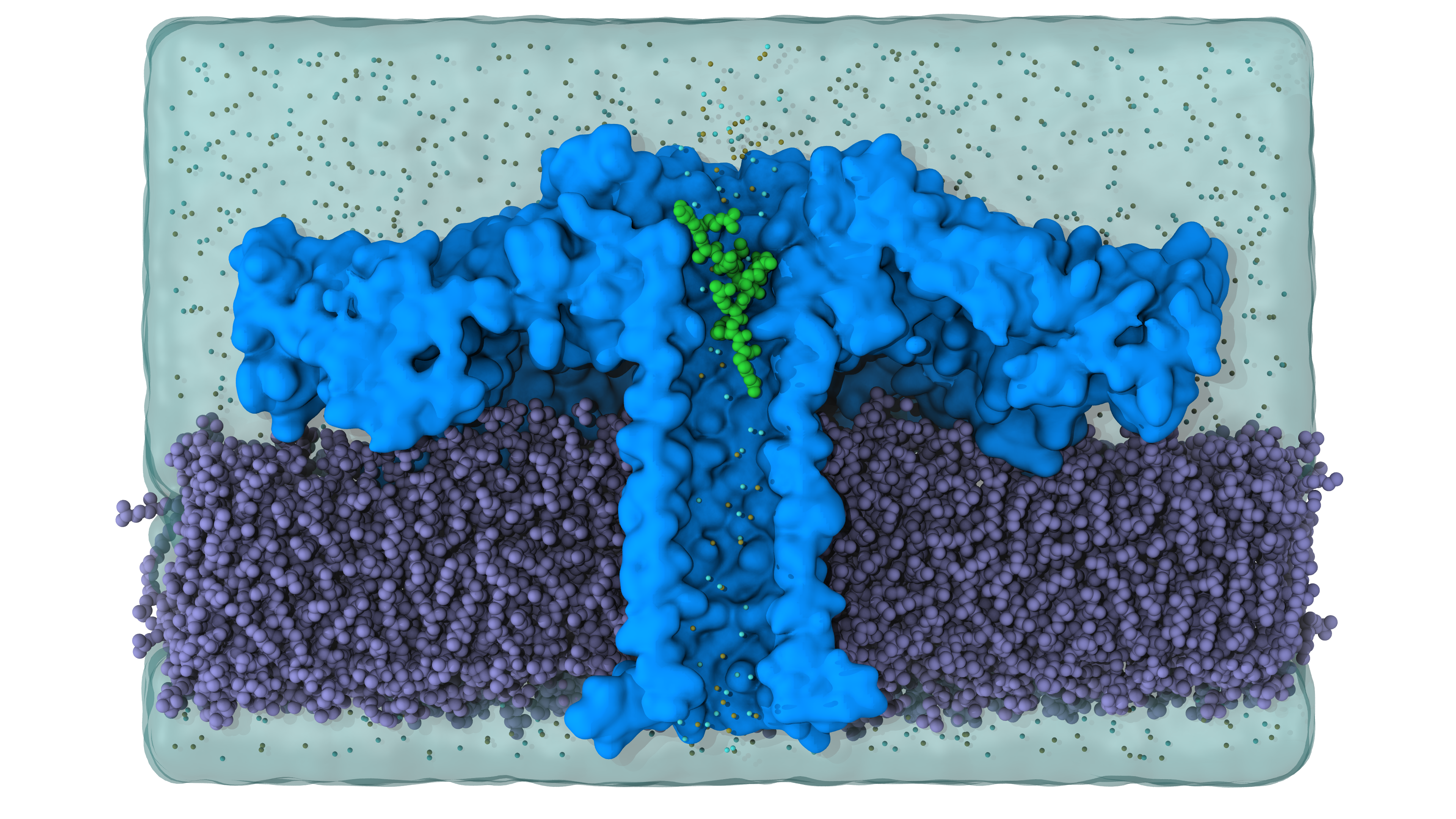}
	\caption{Rendering of an aerolysin nanopore embedded in a DPhPC lipid bilayer with a peptide (green) at the pore's entrance.
		The surrounding fluid contains charged ions at 2M concentration, represented as free spheres in the fluid.}
	\label{fig:peptide-render}
\end{figure*}

\begin{figure*}[ht]
	\centering
	\includegraphics[width=0.9\linewidth]{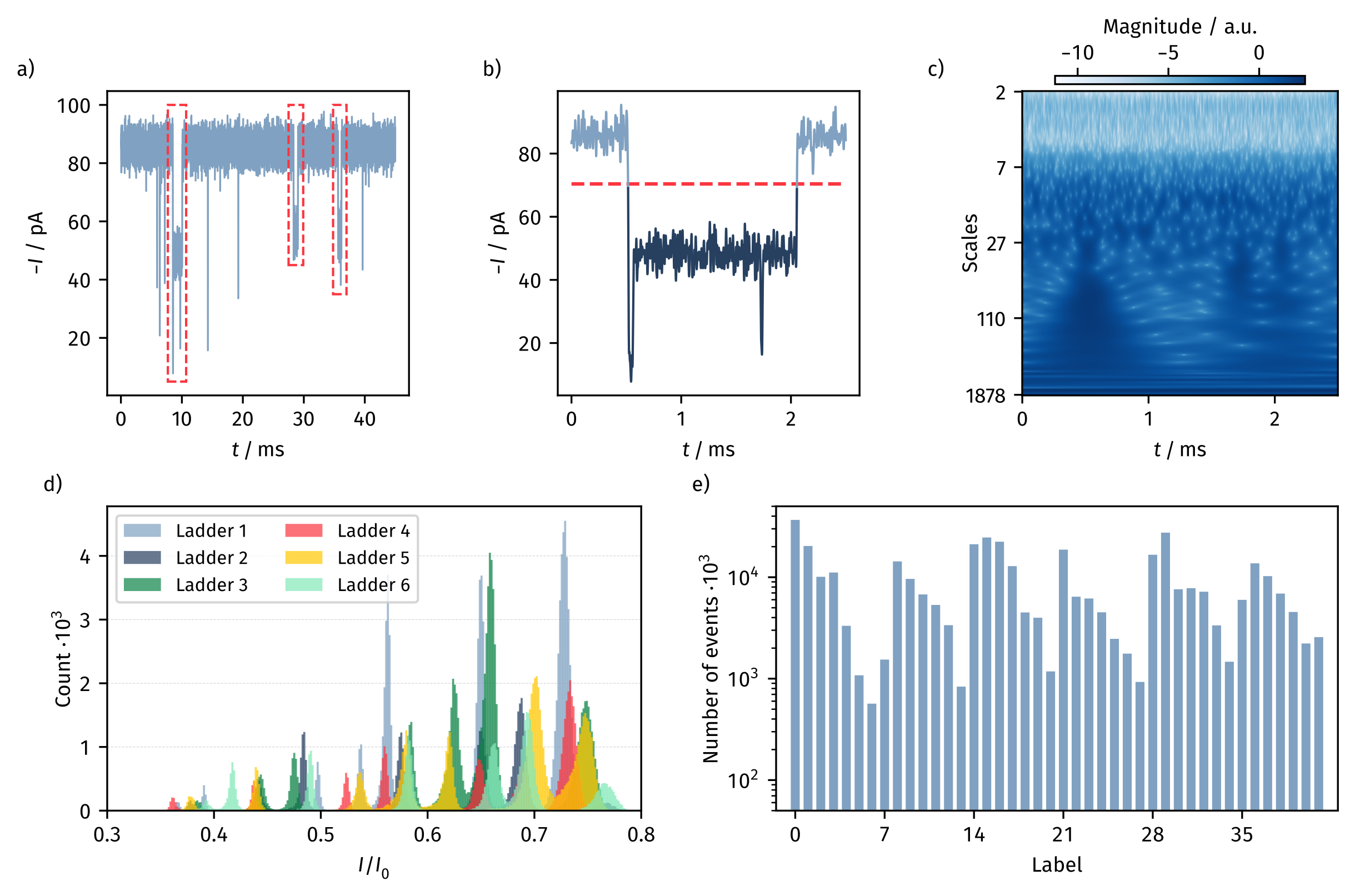}
	\caption{Example data pipeline from experimental readout to wavelet transforms. a) Data is read from an experiment, resulting in a range of events corresponding to peptides entering the pore. b) Events are detected using a standard, three-sigma algorithm. c) A wavelet transform is applied to individual events, resulting in a scaleogram. d) Histograms of the individual peptide ladder experiments. The six ladders were recorded in separate experiments and their histograms are overlaid here for comparison. The extent of the overlap highlights the need for advanced classification algorithms, since a simple mean-current analysis would not suffice to identify single peptides were all of these peptides present in one sample. e) Distribution of extracted, labeled events, split at the level of individual events into training, validation and test sets such that the label imbalance was preserved.}
	\label{fig:data-processing}
\end{figure*}

Initial data was generated by nanopore experiments carried out at the University of Freiburg~\cite{behrends22a,piguet21a}, details of which can be found in the Methods Section.
Figure~\ref{fig:peptide-render} illustrates the physical process, reproduced in simulation, by which a small peptide in a bath of ions enters a pore reconstituted in a membrane.
Experiments such as these generate large current sequences from which individual events must be extracted and labeled.
In Figures~\ref{fig:data-processing} a) and b), the raw data extracted from the experiments, together with an event extracted, are displayed.
The event extraction process is elaborated in the Methods Section.
These extracted events can be used to construct histograms of the relative blockade current of all the peptides available for classification, as outlined in Figure~\ref{fig:data-processing} d), where 42 individual, partially overlapping distributions correspond to the 42 peptides in the solutions.
The fact that these histograms overlap to such a degree represents the limitation of blockade current averages to distinguish between many peptide classes.
Finally, in Figure~\ref{fig:data-processing} e), a histogram of the data quantity per class is outlined to highlight the imbalance of samples between the classes used in training.
This data was further split into test, training and validation as detailed in the Methods Section.
The split is consequently over individual events, and the resulting accuracies quantify generalization to unseen events from the same recordings subject to the same experimental conditions.

\subsection*{Wavelet transformations and scaleogram construction}

Once the data is extracted and labeled, each blockade current (Figure~\ref{fig:data-processing} b) as an example) is converted into a complex-valued scaleogram image via continuous wavelet transform using the ssqueezepy Python package~\citep{muradeli20a}, the result is shown in Figure~\ref{fig:data-processing} c), by taking the absolute value of the transform for each pixel.
Detailed information on the wavelet transformation can be found in the Methods Section. 
The scaleogram contains all of the time and frequency information from the current signal, represented as an image.
For training, the absolute value of the continuous wavelet transform is used, and an element-wise logarithm of the values is applied before resizing in order to improve the numeric stability of the training.

\subsection*{Model training and classification performance}

In order to assess the functionality of image representations for peptide classification, three neural network architectures were chosen due to their ubiquity and state-of-the-art performance on vision tasks.
These networks included a ResNet18~\citep{he15a}, a ResNeXt101~\citep{xie17a}, and a Vision Transformer~\citep{dosovitskiy21a}.
Additional details about the architectures and how they were trained can be found in the Methods Section. 
Final macro, micro and top-10 accuracies produced by each model are shown in Table~\ref{tab:performance}, including a comparison to prior work~\citep{hossbach25a}.
\begin{table*}[ht]
	\caption{Test performance of the classifier models used in this investigation, along with a comparison to prior work~\citep{hossbach25a}.}
	\label{tab:performance}
	\begin{tabular}{@{}ccccc@{}}
		\toprule
		Averaging & ResNet18 & ResNeXt101 & Vision Transformer & catch22~\citep{hossbach25a} \\ \midrule
		Macro     & 82.2 \%  & 80.2 \%    & 81.4 \%            & 73.6 \%                     \\
		Micro     & 82.1 \%  & 79.8 \%    & 81.6 \%            & 73.6 \%                     \\
		Top-10    & 84.7 \%  & 84.8 \%    & 83.8 \%            & 75.8 \%                     \\ \bottomrule
	\end{tabular}
\end{table*}
Here, the macro-averaged accuracy is defined as the arithmetic mean of all per-class accuracies, whereas the micro-averaged accuracy is defined as the accuracy over all samples in the dataset.
For datasets with high data imbalance, the latter is biased towards classes with more samples.
Calculating the accuracy for individual classes and averaging afterwards avoids this bias.
Finally, the top-10 accuracy denotes the average accuracy of the ten classes with the most samples in the dataset.
The models trained on the scaleograms outperform our previous approach of using the catch22 feature set in all shown metrics, with the best-performing ResNet18 exceeding it by $8.6$ percentage points in macro accuracy, $8.5$ in micro accuracy and $8.9$ in top-10 accuracy.
We note that the catch22 values are reproduced from our earlier work~\citep{hossbach25a} and were not re-trained on the identical event split used here.
The comparison is therefore indicative of the benefit of the scaleogram-based pipeline as a whole rather than a controlled ablation of its individual components.
The ResNet18 model achieves the best overall performance, reaching a macro accuracy of over $82\,\%$.
The values reported here are those of the best run of each architecture.
The balanced validation accuracies of the individual runs, one per initialization seed, are listed in the Supplementary Information.
Although this suggests that the ResNet18 architecture is best suited for this task, we consider it likely that the result is limited by the amount of training data available.
This is consistent with the top-10 accuracies of all models being higher than the respective macro accuracies, although we note that the top-10 metric also averages over fewer classes, so this observation is suggestive rather than decisive.
Furthermore, the ResNeXt101 model performs worse than the ResNet18 model, even though the larger model should, in principle, outperform their smaller counterparts given sufficient data~\citep{yang22a}.
Due to experimental constraints on data generation, the networks were only trained on approximately $254,000$ images, far below the standard in larger vision models such as the ImageNet benchmark~\citep{deng09a}.
The results therefore suggest that the larger ResNeXt101 has not reached its limit in performance, and that the models may improve further with more data.

\begin{figure*}[ht]
	\centering
	\includegraphics[width=\linewidth]{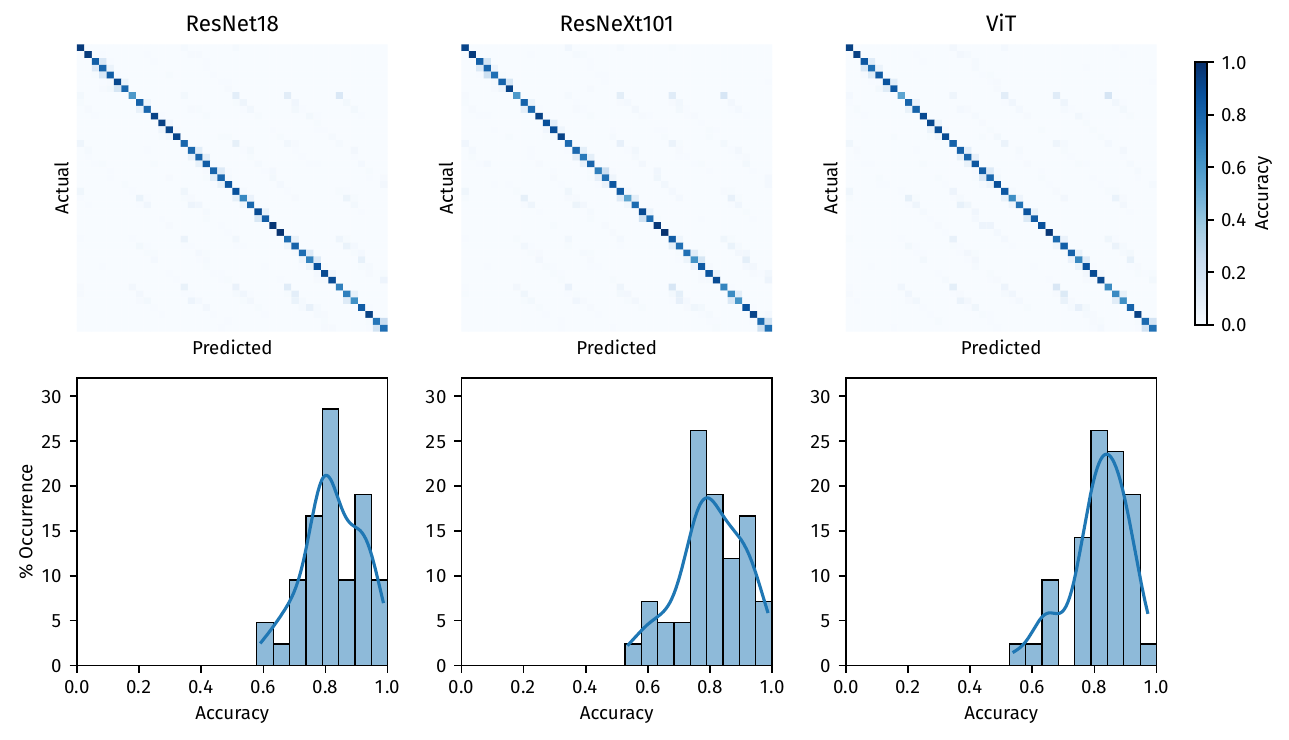}
	\caption{
		Confusion matrices and accuracy distribution for the ResNet18, ResNeXt101 and Vision transformer architectures assessed throughout this work.
		Here, the accuracy denotes the per-class accuracy, where row-wise normalization was performed.
	}
	\label{fig:classification-results}
\end{figure*}

To understand the misclassifications by the models, we plot the confusion matrices computed over the test dataset, as shown in the upper portion of Figure~\ref{fig:classification-results}.
These 42 by 42 matrices are constructed by populating their elements with the test samples, where the row corresponds to the sample's label, and the column is given by the model's prediction.
For a complete list of which peptide corresponds to which label, see Table 1 in the Supplementary Information.
A larger number on the diagonal corresponds to a better performance of the model, whereas off-diagonal elements in the confusion matrices highlight classes that are confused with others.
In the confusion matrices, a periodic pattern in the off-diagonals can be observed.
These elements correspond to misclassifications of peptides with equal-length peptides from other ladders, whose mean blockade current histograms also show partial or complete overlap.
Based on the partial overlap seen in Figure~\ref{fig:data-processing}d) we believe these misclassifications to stem either from the numerical transformation of the currents into scaleograms or due to intrinsic similarities in the peptide dynamics beyond the mean blockade.

Further insights on the subtle differences between the model results is gained by looking at the distribution of macro accuracies of the models, shown in the lower portion of Figure~\ref{fig:classification-results}.
The overall classification performance varies notably across models.
ResNet18 demonstrates the best overall performance, with a narrower distribution of class accuracies that is shifted closer to 1.
In contrast, the ResNeXt101 shows a broader distribution, reflecting a greater variability in the class-wise accuracy.
In terms of minimal class accuracy, differences between models are also evident:
for the particularly challenging class L2AA4, the ResNeXt101 model achieves the lowest accuracy at $53.8\,\%$.
This improves to $54.1\,\%$ with the ViT, and reaches $59.3\,\%$ with the ResNet18 model.
However, the ResNet18's performance is likely related to the limited size of the training dataset, and we expect the more complex models to outperform it with more data.

\subsection*{Importance analysis}

The scaleogram representation allows for a frequency-resolved interpretation of relevant dynamics in the signals as predicted by the trained models.
To realize this, a DeepliftSHAP analysis was performed using the Captum framework~\citep{kokhlikyan20a-pre} and the best-performing ResNet18 architecture.
Further implementation details can be found in the Methods Section. 

\begin{figure*}[h]
	\centering
	\includegraphics{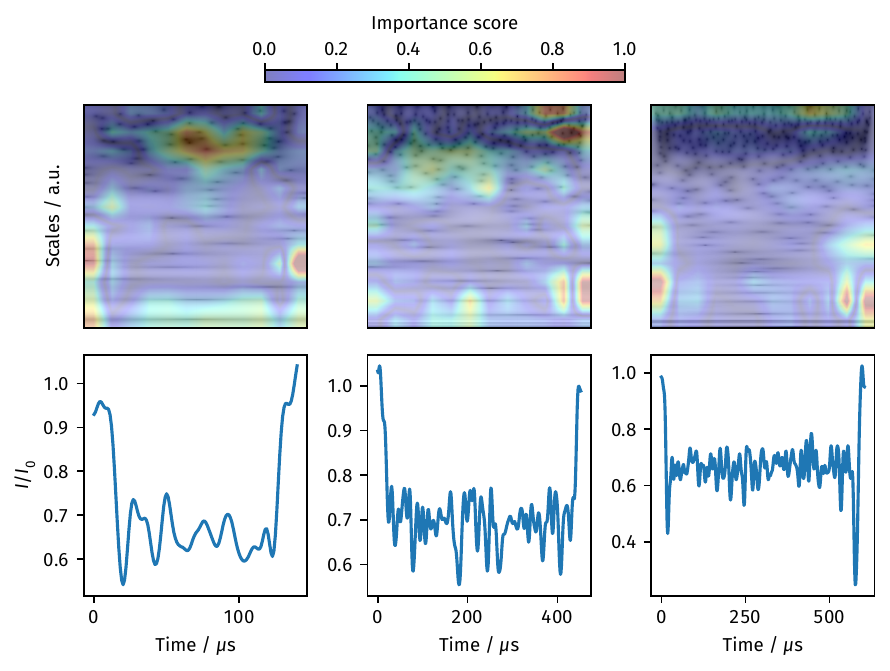}
	\caption{
		Visualization of scaleograms randomly chosen from the test dataset with important regions, based on the importance score calculated using the DeepLiftSHAP algorithm.
		Note that all CWT images have been rescaled to equal size, thereby removing explicit information about the original scale axes.
		The importance scores are broadened for better visualization.
		Attributions are shown for three events drawn at random from the test set, without filtering on whether the model classified them correctly, and were computed against a single reference distribution of open-pore scaleograms (see Methods).
	}
	\label{fig:attentions}
\end{figure*}

Figure~\ref{fig:attentions} shows a visualization of the important regions for three samples drawn at random from the test dataset.
In all three examples, a high importance is assigned to the lower corners of the scaleogram, reflecting rise- and fall-events of the blockade current and thus capturing information about the magnitude of the current blockade.
More surprisingly, a high importance score is also observed in the upper portion of the scaleogram.
One interpretation is that the model recognizes aspects of the high-frequency portions of the signal that are otherwise dominated by noise.
A larger set of $80$ attribution maps, drawn at random from the test set without filtering on whether the model classified the event correctly, is provided in the Supplementary Information.

\subsection*{Pruning and quantization study}

\begin{figure*}[ht]
	\centering
	\includegraphics{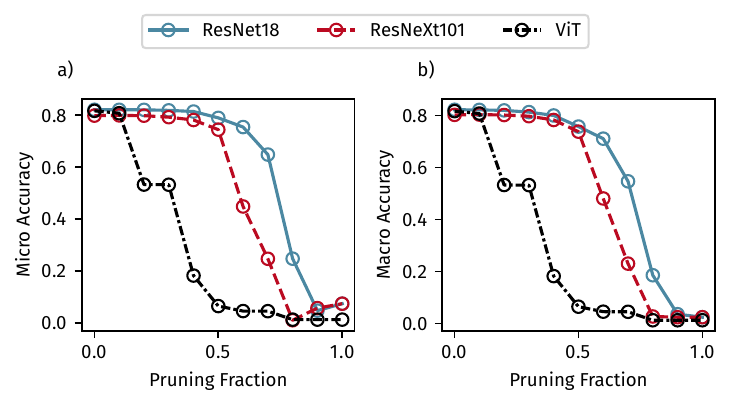}
	\caption{Micro accuracy (a) and Macro accuracy (b) of the models as a function of the pruning fraction.}
	\label{fig:network-pruning}
\end{figure*}

A critical aspect for the use of the classification algorithm in diagnostics is the ability to deploy the trained models locally.
Allowing for local deployment will improve data security and provide a means for offline functionality, essential in many point-of-care applications~\citep{ratinho25a}.
However, as these models become larger and handle more classes, the problem of deploying them into a device will be challenging to solve.
To this end, we explore the performance of the three introduced models after undergoing a model-compression pipeline.
In realistic deployment scenarios, this pipeline consists first of methods to reduce the size and complexity of the networks via weight pruning~\citep{blalock20a, liang21a, frankle19a} and quantization~\citep{nagel21a, choi17a}.
Weight pruning involves removing network parameters based on their magnitude and is often characterized by a removal fraction, the percentage of parameters that were removed.
Quantization reduces the precision of parameter values by mapping them to a smaller set of discrete levels in a lower-bit representation.
For pruning, both structured and unstructured weight pruning were studied; however, only the global unstructured pruning produced usable results.
Details on the implementation can be found in the Methods Section. 
To measure the effect of pruning, the micro and macro accuracy scores are computed to demonstrate how the models' performance deteriorates after pruning.
Figure~\ref{fig:network-pruning} outlines the results of this investigation.

Micro and macro accuracy reveal identical trends, corroborating degradation to be class-independent.
The ResNet18 model appears to be the most resilient to pruning, retaining its accuracy up to a sparsity of $50\,\%$, that is, when half of its weights have been set to zero.
The ResNeXt101 shows similar resilience, whereas the ViT architecture is far more sensitive.
While these results are promising, it should be noted that in well-trained models, pruning up to $90\,\%$ can be achieved without significant degradation of model performance~\citep{frankle19a}.
This result is merely a demonstration that the models can be improved and optimized for hardware deployment once trained on sufficient data, particularly in the case of the ViT models, where many alternative pruning and model reduction approaches are being developed~\citep{zhang22a}.
However, it also shows a limitation of the current models in that they are likely not sufficiently well trained to have constructed stable latent space representations and therefore, are highly sensitive to changes in their structure.
Remedies to this problem can include adding dropout-style training into the optimization procedure, but would likely also be aided by sufficient training data.

For quantization, we used Post Training Static Quantization for the ResNet18 and ResNeXt101 models, and Dynamic Quantization for the Vision Transformer, using PyTorch's Quantization framework~\citep{paszke19a}.
The ResNet18 and ResNeXt101 models were calibrated on the validation dataset, and all quantized models were benchmarked on the test dataset.
The results are shown in Table~\ref{tab:quantization}.
Among all three models, the ResNet18 achieves the most effective compression with a minimal drop in macro accuracy of $1.2$ percentage points.
The ResNeXt101, while also achieving a comparable reduction in size, suffers from a significant reduction in performance.
In contrast, the Vision Transformer shows a more modest decrease in size paired by a moderate performance loss.
These differences in performance degradation for both the model pruning and quantization can mainly be attributed to the differences in architecture and parameter redundancy due to the size of the models.
With more data, as well as more sophisticated pruning and quantization methods, we are confident that the ResNeXt101 model and the Vision Transformer may outperform the ResNet18 architecture.
The model sizes reported in Table~\ref{tab:quantization} refer to the storage footprint of the serialized parameters.

\begin{table*}[ht]
	\centering
	\caption{
		Results of the model quantization.
		FP32 denotes the single-precision floating point format, whereas I8 denotes the signed 8-bit integer format.
		The bold numbers in the model size column indicate the compression factors of the quantized models.
	}
	\label{tab:quantization}
	\renewcommand{\arraystretch}{1.15}
	\setlength{\tabcolsep}{11pt}
	\medskip
	\begin{tabular}{lcllcc}
		\toprule
		Model                       & Params ($10^6$)         & Precision & Size (MB)                      & Macro (\%) & Micro (\%)                      \\
		\midrule
		\multirow{2}{*}{ResNet18}   & \multirow{2}{*}{$11.2$} & FP32      & $44.8$                         & $82.2\,\%$ & $82.1\,\%$                      \\
		                            &                         & I8        & $11.3$ ($\sim \mathbf{3.96x}$) & $81.0\,\%$ & $80.1\,\%$                      \\ \cmidrule{1-6}
		\multirow{2}{*}{ResNeXt101} & \multirow{2}{*}{$81.5$} & FP32      & 327                            & $80.2\,\%$ & $79.8\,\%$      \rule{0pt}{2ex} \\
		                            &                         & I8        & $83.5$ ($\sim \mathbf{3.92x}$) & $71.2\,\%$ & $69.3\,\%$                      \\ \cmidrule{1-6}
		\multirow{2}{*}{ViT}        & \multirow{2}{*}{$9.63$} & FP32      & $38.58$                        & $81.4\,\%$ & $81.6\,\%$      \rule{0pt}{2ex} \\
		                            &                         & I8        & $19.5$ ($\sim \mathbf{1.97x}$) & $79.8\,\%$ & $80.2\,\%$                      \\
		\bottomrule
	\end{tabular}
\end{table*}

\section*{Discussion}
In this work, we have demonstrated that image-based machine learning models—including both convolutional neural networks and transformer architectures—combined with scaleogram transformations can achieve up to $82.2\,\%$ accuracy on held-out events in discriminating among 42 distinct peptide classes, recorded as six separate peptide ladders using a biological nanopore. Notably, many of these peptides share highly similar sizes and chemical properties, resulting in minimal or even imperceptible differences in their raw current signals.

During our study, we tested several architectures: ResNet18, a larger ResNeXt101, and a vision transformer. ResNet18 produced the best results after training.
The accuracy reached by all models exceeded that of the prior approach using statistical moments and catch22 features of the current traces~\citep{hossbach25a}.
A plausible explanation for this improvement is the explicit incorporation of signal dynamics into the classification process, as the use of scaleograms enables the models to capture joint time–frequency patterns that are not accessible with the static statistical descriptors previously used.
In contrast to the commonly observed improvement in performance with larger models, the substantially larger ResNeXt101 did not outperform the smaller architectures considered here.
This result departs from empirical scaling behavior reported in language~\citep{kaplan20a-pre} and computer vision~\citep{dosovitskiy21a}, which we tentatively attribute to being data-bound by the limited size of the training dataset.
This assumption is derived from comparative image classification studies in which these models are trained with substantially larger amounts of data~\citep{yu22a,ridnik21a}.
Consequently, we expect the amount of training data available to be a key factor in further improving the performance and extending this peptide classification approach to clinically relevant settings.

Next, in order to further analyze the performance of the models, we computed importance scores for the regions of the scaleogram images using the DeepLiftSHAP algorithm~\citep{lundberg17a}.
This allows a better understanding of how specific features in the image impact the model's predictions, thus highlighting important regions.
Our results indicate that the rise and fall-dynamics observed along the lower edges of the scaleogram are assigned to high importance scores.
This may be because these dynamics capture information about the depth of the blockade current and could, by extension, also encode dwell-time information that is otherwise lost during the rescaling of the scaleogram.
We also found that the noise-dominated regions of the images, attributed to convolutions with daughter wavelets with low scales and consequently high (local) frequencies, received high importance scores.
One possible explanation is that relevant processes take place on short time-scales at this resolution, which would motivate the use of high-bandwidth measurements or of computational studies able to separate signal components from noise-affected data.
Importance in this region could in principle also arise from instrumental noise or from interpolation artifacts introduced when the scaleograms are resized to a common resolution.
We consider this less likely, as the reference distribution used for computing the attributions consists of scaleograms of the open pore current taken from all measurement conditions and processed identically, so that contributions common to the recordings and to the resizing operation are largely averaged out.
In addition, high importance scores were also assigned to an additional blockade state with amplitudes distinct from the main current level of an event.
This additional state, known as the return/deep-level, has been shown~\citep{boukhet17a} and thoroughly explained~\citep{boukhet18a,ensslen24a} previously.
In essence, aerolysin events reveal a two-level nature with a transient deep state that gives information about the position of the peptide along the pore’s channel.
The high importance scores on the lower edges of the corresponding scaleograms are consistent with the model detecting these deep levels and using them to form its predictions.
This suggests that, together with other signal features, the deep-level states play a role in the discrimination of the peptides, which is in line with previous studies focusing on the dynamic features in events involving single-stranded DNA~\citep{wang25a}.

As a step towards the development of offline, possibly embedded devices, we performed model compression studies.
Whether used in point-of-care settings or to comply with data privacy regulations, nanopore-based diagnostics may require easily portable and offline-capable devices during deployment.
Such devices would necessitate the compression of the models as much as possible to improve power consumption and latency.
Our results demonstrated that up to $50\,\%$ of the model weights could be set to zero without a decrease in precision.
Furthermore, quantization achieved up to a $3.96\times$ reduction in the stored model size with little to no degradation in the most robust architecture.
These results characterize how much redundancy the trained models contain and how far their numerical precision can be reduced.

Based on our findings, we identify several directions for future research. 
First, data set constraints remain a critical factor in improving model performance and robustness.
Due to the stochastic nature of the blockade mechanism and variations in experimental conditions, such as ambient temperature, humidity, and electrode state, the development of robust peptide classification methods requires large and diverse training datasets.
In addition, increasing the number of peptides will be an important step towards sampling out the intrinsic data manifold, which would not only improve the robustness and general performance of the models but also be crucial for future diagnostics applications.
In this context, synthetic data generation represents a promising candidate for increasing the size of the data.
This could include in-silico simulations of blockade events to generate blockade event data, or by utilizing generative models to synthesize current traces.
Augmentations of already existing traces, such as Brownian motion augmentation~\citep{kipen25a}, could also significantly increase the size of the data set.

Finally, training-aware quantization deserves further investigation.
Quantization is often applied post hoc, but incorporating quantization constraints directly during training could lead to more efficient models without sacrificing accuracy~\citep{nagel22a}.
This is especially relevant for deployment in resource-constrained environments, where memory footprint and inference latency are important considerations.

\section*{Conclusions}
We have shown that converting nanopore resistive-pulse signals into wavelet scaleogram images, combined with image-based deep learning models, enables the discrimination of 42 closely related peptides with up to $82.2\,\%$ accuracy on held-out events, the highest accuracy reported to date for this dataset.
The improvement over descriptor-based approaches is consistent with the explicit encoding of joint time–frequency signal dynamics.
Our importance analysis suggests that the models make use of physically interpretable features such as the deep-level blockade states.
The models further tolerate substantial pruning and quantization with little loss of accuracy, which is a prerequisite for deployment on portable, offline-capable hardware.
Together, these results are a step towards automated peptide classification in nanopore recordings.
Extending them towards protein identification in clinical samples will require evaluation on independently acquired measurement conditions, on blinded mixtures and in biological matrices, as well as benchmarking on target devices, with the size and diversity of the available training data identified as the key lever for further gains.

\subsection*{Availability of data and materials}
Upon publication, the data will be made available upon reasonable request to the relevant authors.
All scripts used will be published to \url{https://doi.org/10.18419/DARUS-5740}.

\subsection*{Competing interests}
The authors declare the following competing financial interests: The University of Stuttgart has filed a patent application with patent number DE 10 2024 121 850.9 for the methods introduced in this work.

\subsection*{Acknowledgements}
The authors acknowledge Michel Mom for rendering the final version of the nanopore figure.
The authors thank the International Max Planck Research School for Intelligent Systems (IMPRS-IS) for supporting Julian Ho{\ss}bach.
This project was funded by the BMBFTR through the nanodiagBW project, grant No. 03ZU1208AM.
C.H. acknowledges further financial support from the German Funding Agency (Deutsche Forschungsgemeinschaft DFG) under Germany's Excellence Strategy EXC 2075-390740016 and the SPP 2363 - "Utilization and Development of Machine Learning for Molecular Applications – Molecular Machine Learning", Project No. 497249646. \\
Computations were performed on the ICP Compute Cluster, Grant No. INST 41/1148-1 Holm 492175459, and through the INST 35/1597-1 FUGG grant. \\
The authors acknowledge support by the High Performance and Cloud Computing Group at the Zentrum für Datenverarbeitung of the University of Tübingen, the state of Baden-Württemberg through bwHPC and the German Research Foundation (DFG) for funding under "Project number 455787709" (bwForCluster BinAC 2).

\subsection*{Authors' contributions}
J.H. conducted model training and analysis, contributed to manuscript writing, and performed editing.
S.T. came up with the approach used in the paper, conducted model training and analysis, drafted the manuscript, and performed editing and submission.
S.K. conducted model training and analysis and contributed to writing the manuscript.
T.E. generated the data used for training and contributed to manuscript writing.
J.C.B. supervised the generation of the data.
C.H. supervised the investigation and provided feedback.
All authors read the manuscript and provided editing feedback.

\newpage
\bibliography{bibliography}

@article{barker23a,
  title = {An Inflection Point in Cancer Protein Biomarkers: What was and What's
           Next},
  journal = {Molecular \& Cellular Proteomics},
  volume = {22},
  number = {7},
  pages = {100569},
  year = {2023},
  issn = {1535-9476},
  doi = {https://doi.org/10.1016/j.mcpro.2023.100569},
  url = {https://www.sciencedirect.com/science/article/pii/S1535947623000804},
  author = {Anna D. Barker and Mario M. Alba and Parag Mallick and David B. Agus
            and Jerry S.H. Lee},
}

@article{motone24a,
  title = {Multi-Pass, Single-Molecule Nanopore Reading of Long Protein Strands},
  author = {Motone, Keisuke and {Kontogiorgos-Heintz}, Daphne and Wee, Jasmine
            and Kurihara, Kyoko and Yang, Sangbeom and Roote, Gwendolin and Fox,
            Oren E. and Fang, Yishu and Queen, Melissa and Tolhurst, Mattias and
            Cardozo, Nicolas and Jain, Miten and Nivala, Jeff},
  year = 2024,
  month = sep,
  journal = {Nature},
  volume = {633},
  number = {8030},
  pages = {662--669},
  issn = {0028-0836, 1476-4687},
  doi = {10.1038/s41586-024-07935-7},
  urldate = {2025-01-21},
  langid = {english},
}

@article{wang25a,
  title = {Dynamic {{Features Driven}} by {{Stochastic Collisions}} in a {{Nanopore}} for {{Precise Single-Molecule Identification}}},
  author = {Wang, Jia and Liu, Shao-Chuang and Hu, Zheng-Li and Ying, Yi-Lun and Long, Yi-Tao},
  year = 2025,
  month = jan,
  journal = {Journal of the American Chemical Society},
  volume = {147},
  number = {2},
  pages = {1781--1791},
  issn = {0002-7863, 1520-5126},
  doi = {10.1021/jacs.4c13664},
  urldate = {2025-03-31},
  langid = {english}
}

@article{wang24a,
  title = {Unambiguous Discrimination of All 20 Proteinogenic Amino Acids and
           Their Modifications by Nanopore},
  author = {Wang, Kefan and Zhang, Shanyu and Zhou, Xiao and Yang, Xian and Li,
            Xinyue and Wang, Yuqin and Fan, Pingping and Xiao, Yunqi and Sun, Wen
            and Zhang, Panke and Li, Wenfei and Huang, Shuo},
  year = 2024,
  month = jan,
  journal = {Nature Methods},
  volume = {21},
  number = {1},
  pages = {92--101},
  publisher = {Nature Publishing Group},
  issn = {1548-7105},
  doi = {10.1038/s41592-023-02021-8},
  urldate = {2024-08-19},
  copyright = {2023 The Author(s), under exclusive licence to Springer Nature
               America, Inc.},
  langid = {english},
}

@article{kipen25a,
  title = {Brownian Motion Data Augmentation: A Method to Push Neural Network
           Performance on Nanopore Sensors},
  shorttitle = {Brownian Motion Data Augmentation},
  author = {Kipen, Javier and Jald{\'e}n, Joakim},
  editor = {Wren, Jonathan},
  year = 2025,
  month = jun,
  journal = {Bioinformatics},
  volume = {41},
  number = {6},
  pages = {btaf323},
  issn = {1367-4811},
  doi = {10.1093/bioinformatics/btaf323},
  urldate = {2025-08-04},
  copyright = {https://creativecommons.org/licenses/by/4.0/},
  langid = {english},
}

@inproceedings{dosovitskiy21a,
  title = {An Image Is Worth 16x16 Words: {{Transformers}} for Image Recognition
           at Scale},
  booktitle = {International Conference on Learning Representations},
  author = {Dosovitskiy, Alexey and Beyer, Lucas and Kolesnikov, Alexander and
            Weissenborn, Dirk and Zhai, Xiaohua and Unterthiner, Thomas and
            Dehghani, Mostafa and Minderer, Matthias and Heigold, Georg and Gelly
            , Sylvain and Uszkoreit, Jakob and Houlsby, Neil},
  year = 2021,
}

@misc{kaplan20a-pre,
  title = {Scaling {{Laws}} for {{Neural Language Models}}},
  author = {Kaplan, Jared and McCandlish, Sam and Henighan, Tom and Brown, Tom
            B. and Chess, Benjamin and Child, Rewon and Gray, Scott and Radford,
            Alec and Wu, Jeffrey and Amodei, Dario},
  year = 2020,
  publisher = {arXiv},
  doi = {10.48550/ARXIV.2001.08361},
  urldate = {2026-01-12},
  copyright = {arXiv.org perpetual, non-exclusive license},
}

@article{robertson07a,
  title = {Single-Molecule Mass Spectrometry in Solution Using a Solitary
           Nanopore},
  author = {Robertson, Joseph W. F. and Rodrigues, Claudio G. and Stanford,
            Vincent M. and Rubinson, Kenneth A. and Krasilnikov, Oleg V. and
            Kasianowicz, John J.},
  date = {2007-05-15},
  journaltitle = {Proceedings of the National Academy of Sciences},
  shortjournal = {Proc. Natl. Acad. Sci. U.S.A.},
  volume = {104},
  number = {20},
  pages = {8207--8211},
  issn = {0027-8424, 1091-6490},
  doi = {10.1073/pnas.0611085104},
  url = {https://pnas.org/doi/full/10.1073/pnas.0611085104},
  urldate = {2026-01-08},
  langid = {english},
}

@article{ensslen22a,
  title = {Resolving {{Isomeric Posttranslational Modifications Using}} a {{
           Biological Nanopore}} as a {{Sensor}} of {{Molecular Shape}}},
  author = {Ensslen, Tobias and Sarthak, Kumar and Aksimentiev, Aleksei and
            Behrends, Jan C.},
  date = {2022-09-07},
  journaltitle = {Journal of the American Chemical Society},
  shortjournal = {J. Am. Chem. Soc.},
  volume = {144},
  number = {35},
  pages = {16060--16068},
  publisher = {American Chemical Society},
  issn = {0002-7863},
  doi = {10.1021/jacs.2c06211},
  url = {https://doi.org/10.1021/jacs.2c06211},
  urldate = {2023-03-22},
}

@article{kasianowicz96a,
  title = {Characterization of Individual Polynucleotide Molecules Using a
           Membrane Channel},
  author = {Kasianowicz, John J. and Brandin, Eric and Branton, Daniel and
            Deamer, David W.},
  date = {1996-11-26},
  journaltitle = {Proceedings of the National Academy of Sciences},
  shortjournal = {Proc. Natl. Acad. Sci. U.S.A.},
  volume = {93},
  number = {24},
  pages = {13770--13773},
  publisher = {National Acad Sciences},
  issn = {0027-8424, 1091-6490},
  doi = {10.1073/pnas.93.24.13770},
  url = {https://pnas.org/doi/full/10.1073/pnas.93.24.13770},
  urldate = {2026-01-08},
  langid = {english},
}

@article{whiteaker14a,
  author = {Whiteaker, Jeffrey R. and Halusa, Goran N. and Hoofnagle, Andrew N.
            and Sharma, Vagisha and MacLean, Brendan and Yan, Ping and Wrobel,
            John A. and Kennedy, Jacob and Mani, D. R. and Zimmerman, Lisa J. and
            Meyer, Matthew R. and Mesri, Mehdi and Rodriguez, Henry and
            Abbatiello, Susan E. and Boja, Emily and Carr, Steven A. and Chan,
            Daniel W. and Chen, Xian and Chen, Jing and Davies, Sherri R. and
            Ellis, Matthew J. C. and Feny{\"o}, David and Hiltke, Tara and
            Ketchum, Karen A. and Kinsinger, Chris and Kuhn, Eric and Liebler,
            Daniel C. and Lin, De and Liu, Tao and Loss, Michael and MacCoss,
            Michael J. and Qian, Wei-Jun and Rivers, Robert and Rodland, Karin D.
            and Ruggles, Kelly V. and Scott, Mitchell G. and Smith, Richard D.
            and Thomas, Stefani and Townsend, R. Reid and Whiteley, Gordon and Wu
            , Chaochao and Zhang, Hui and Zhang, Zhen and Paulovich, Amanda G.
            and the Clinical Proteomic Tumor Analysis Consortium (CPTAC)},
  title = {CPTAC Assay Portal: a repository of targeted proteomic assays},
  journal = {Nature Methods},
  year = {2014},
  month = {7},
  day = {01},
  volume = {11},
  number = {7},
  pages = {703-704},
  issn = {1548-7105},
  doi = {10.1038/nmeth.3002},
  url = {https://doi.org/10.1038/nmeth.3002},
}

@article{nagaraju22a,
  author = {Nagaraju, M. and Chawla, Priyanka and Kumar, Neeraj},
  title = {Performance improvement of Deep Learning Models using image
           augmentation techniques},
  journal = {Multimedia Tools and Applications},
  year = {2022},
  month = {Mar},
  day = {01},
  volume = {81},
  number = {7},
  pages = {9177-9200},
  issn = {1573-7721},
  doi = {10.1007/s11042-021-11869-x},
  url = {https://doi.org/10.1007/s11042-021-11869-x},
}

@misc{yang22a,
  title = {Tensor Programs V: Tuning Large Neural Networks via Zero-Shot
           Hyperparameter Transfer},
  author = {Greg Yang and Edward J. Hu and Igor Babuschkin and Szymon Sidor and
            Xiaodong Liu and David Farhi and Nick Ryder and Jakub Pachocki and
            Weizhu Chen and Jianfeng Gao},
  year = {2022},
  eprint = {2203.03466},
  archivePrefix = {arXiv},
  primaryClass = {cs.LG},
  url = {https://arxiv.org/abs/2203.03466},
}

@article{son19a,
  author = {Son, Minsoo and Kim, Hyunsoo and Yeo, Injoon and Kim, Yoseop and
            Sohn, Areum and Kim, Youngsoo},
  title = {Method Validation by CPTAC Guidelines for Multi-protein Marker Assays
           Using Multiple Reaction Monitoring-mass Spectrometry},
  journal = {Biotechnology and Bioprocess Engineering},
  year = {2019},
  month = {3},
  day = {01},
  volume = {24},
  number = {2},
  pages = {343-358},
  issn = {1976-3816},
  doi = {10.1007/s12257-018-0454-7},
  url = {https://doi.org/10.1007/s12257-018-0454-7},
}

@article{rudnik16a,
  author = {Rudnick, Paul A. and Markey, Sanford P. and Roth, Jeri and Mirokhin,
            Yuri and Yan, Xinjian and Tchekhovskoi, Dmitrii V. and Edwards,
            Nathan J. and Thangudu, Ratna R. and Ketchum, Karen A. and Kinsinger,
            Christopher R. and Mesri, Mehdi and Rodriguez, Henry and Stein,
            Stephen E.},
  title = {A Description of the Clinical Proteomic Tumor Analysis Consortium
           (CPTAC) Common Data Analysis Pipeline},
  journal = {Journal of Proteome Research},
  volume = {15},
  number = {3},
  pages = {1023-1032},
  year = {2016},
  doi = {10.1021/acs.jproteome.5b01091},
  note = {PMID: 26860878},
  URL = {https://doi.org/10.1021/acs.jproteome.5b01091},
  eprint = {https://doi.org/10.1021/acs.jproteome.5b01091},
}

@article{papier24a,
  author = {Papier, Keren and Atkins, Joshua R. and Tong, Tammy Y. N. and
            Gaitskell, Kezia and Desai, Trishna and Ogamba, Chibuzor F. and
            Parsaeian, Mahboubeh and Reeves, Gillian K. and Mills, Ian G. and Key
            , Tim J. and Smith-Byrne, Karl and Travis, Ruth C.},
  title = {Identifying proteomic risk factors for cancer using prospective and
           exome analyses of 1463 circulating proteins and risk of 19 cancers in
           the UK Biobank},
  journal = {Nature Communications},
  year = {2024},
  month = {May},
  day = {15},
  volume = {15},
  number = {1},
  pages = {4010},
  issn = {2041-1723},
  doi = {10.1038/s41467-024-48017-6},
  url = {https://doi.org/10.1038/s41467-024-48017-6},
}

@article{cox11a,
  author = "Cox, Jürgen and Mann, Matthias",
  title = "Quantitative, High-Resolution Proteomics for Data-Driven Systems
           Biology",
  journal = "Annual Review of Biochemistry",
  year = "2011",
  volume = "80",
  number = "Volume 80, 2011",
  pages = "273-299",
  doi = "https://doi.org/10.1146/annurev-biochem-061308-093216",
  url = "
         https://www.annualreviews.org/content/journals/10.1146/annurev-biochem-061308-093216
         ",
  publisher = "Annual Reviews",
  issn = "1545-4509",
  type = "Journal Article",
}

@article{nowakowski14a,
  author = {Nowakowski, Andrew B and Wobig, William J and Petering, David H},
  title = "{Native SDS-PAGE: high resolution electrophoretic separation of
           proteins with retention of native properties including bound metal
           ions}",
  journal = {Metallomics},
  volume = {6},
  number = {5},
  pages = {1068-1078},
  year = {2014},
  month = {03},
  issn = {1756-5901},
  doi = {10.1039/c4mt00033a},
  url = {https://doi.org/10.1039/c4mt00033a},
  eprint = {
            https://academic.oup.com/metallomics/article-pdf/6/5/1068/34772691/c4mt00033a.pdf
            },
}

@article{wang21a,
  author = {Wang, Yunhao and Zhao, Yue and Bollas, Audrey and Wang, Yuru and Au,
            Kin Fai},
  title = {Nanopore sequencing technology, bioinformatics and applications},
  journal = {Nature Biotechnology},
  year = {2021},
  month = {Nov},
  day = {01},
  volume = {39},
  number = {11},
  pages = {1348-1365},
  issn = {1546-1696},
  doi = {10.1038/s41587-021-01108-x},
  url = {https://doi.org/10.1038/s41587-021-01108-x},
}

@article{dekker07a,
  author = {Dekker, Cees},
  title = {Solid-state nanopores},
  journal = {Nature Nanotechnology},
  year = {2007},
  month = {Apr},
  day = {01},
  volume = {2},
  number = {4},
  pages = {209-215},
  issn = {1748-3395},
  doi = {10.1038/nnano.2007.27},
  url = {https://doi.org/10.1038/nnano.2007.27},
}

@article{cao16a,
  author = {Cao, Chan and Ying, Yi-Lun and Hu, Zheng-Li and Liao, Dong-Fang and
            Tian, He and Long, Yi-Tao},
  title = {Discrimination of oligonucleotides of different lengths with a
           wild-type aerolysin nanopore},
  journal = {Nature Nanotechnology},
  year = {2016},
  month = {Aug},
  day = {01},
  volume = {11},
  number = {8},
  pages = {713-718},
  issn = {1748-3395},
  doi = {10.1038/nnano.2016.66},
  url = {https://doi.org/10.1038/nnano.2016.66},
}

@article{schneider12a,
  author = {Schneider, Gr{\'e}gory F. and Dekker, Cees},
  title = {DNA sequencing with nanopores},
  journal = {Nature Biotechnology},
  year = {2012},
  month = {Apr},
  day = {01},
  volume = {30},
  number = {4},
  pages = {326-328},
  issn = {1546-1696},
  doi = {10.1038/nbt.2181},
  url = {https://doi.org/10.1038/nbt.2181},
}

@article{ying22a,
  author = {Ying, Yi-Lun and Hu, Zheng-Li and Zhang, Shengli and Qing, Yujia and
            Fragasso, Alessio and Maglia, Giovanni and Meller, Amit and Bayley,
            Hagan and Dekker, Cees and Long, Yi-Tao},
  title = {Nanopore-based technologies beyond DNA sequencing},
  journal = {Nature Nanotechnology},
  year = {2022},
  month = {Nov},
  day = {01},
  volume = {17},
  number = {11},
  pages = {1136-1146},
  issn = {1748-3395},
  doi = {10.1038/s41565-022-01193-2},
  url = {https://doi.org/10.1038/s41565-022-01193-2},
}

@article{winston12a,
  title = {DNA Base-Calling from a Nanopore Using a Viterbi Algorithm},
  journal = {Biophysical Journal},
  volume = {102},
  number = {10},
  pages = {L37-L39},
  year = {2012},
  issn = {0006-3495},
  doi = {https://doi.org/10.1016/j.bpj.2012.04.009},
  url = {https://www.sciencedirect.com/science/article/pii/S0006349512004511},
  author = {Winston Timp and Jeffrey Comer and Aleksei Aksimentiev},
}

@article{neumann22,
  author = {Neumann, Don and Reddy, Anireddy S. N. and Ben-Hur, Asa},
  title = {RODAN: a fully convolutional architecture for basecalling nanopore
           RNA sequencing data},
  journal = {BMC Bioinformatics},
  year = {2022},
  month = {Apr},
  day = {20},
  volume = {23},
  number = {1},
  pages = {142},
  issn = {1471-2105},
  doi = {10.1186/s12859-022-04686-y},
  url = {https://doi.org/10.1186/s12859-022-04686-y},
}

@article{boza17a,
  doi = {10.1371/journal.pone.0178751},
  author = {Boža, Vladimír AND Brejová, Broňa AND Vinař, Tomáš},
  journal = {PLOS ONE},
  publisher = {Public Library of Science},
  title = {DeepNano: Deep recurrent neural networks for base calling in MinION
           nanopore reads},
  year = {2017},
  month = {06},
  volume = {12},
  url = {https://doi.org/10.1371/journal.pone.0178751},
  pages = {1-13},
  number = {6},
}

@article{carral21a,
  author = {D{\'i'}az Carral, {\'A'}ngel and Ostertag, Magnus and Fyta, Maria},
  title = {Deep learning for nanopore ionic current blockades},
  journal = {The Journal of Chemical Physics},
  volume = {154},
  number = {4},
  pages = {044111},
  year = {2021},
  month = {01},
  issn = {0021-9606},
  doi = {10.1063/5.0037938},
  url = {https://doi.org/10.1063/5.0037938},
  eprint = {
            https://pubs.aip.org/aip/jcp/article-pdf/doi/10.1063/5.0037938/13943106/044111
            \_1\_online.pdf},
}

@misc{nagel21a,
  title = {A White Paper on Neural Network Quantization},
  author = {Markus Nagel and Marios Fournarakis and Rana Ali Amjad and Yelysei
            Bondarenko and Mart van Baalen and Tijmen Blankevoort},
  year = {2021},
  eprint = {2106.08295},
  archivePrefix = {arXiv},
  primaryClass = {cs.LG},
  url = {https://arxiv.org/abs/2106.08295},
}

@misc{choi17a,
  title = {Towards the Limit of Network Quantization},
  author = {Yoojin Choi and Mostafa El-Khamy and Jungwon Lee},
  year = {2017},
  eprint = {1612.01543},
  archivePrefix = {arXiv},
  primaryClass = {cs.CV},
  url = {https://arxiv.org/abs/1612.01543},
}

@misc{zhang22a,
  title = {PLATON: Pruning Large Transformer Models with Upper Confidence Bound
           of Weight Importance},
  author = {Qingru Zhang and Simiao Zuo and Chen Liang and Alexander Bukharin
            and Pengcheng He and Weizhu Chen and Tuo Zhao},
  year = {2022},
  eprint = {2206.12562},
  archivePrefix = {arXiv},
  primaryClass = {cs.LG},
  url = {https://arxiv.org/abs/2206.12562},
}

@misc{frankle19a,
  title = {The Lottery Ticket Hypothesis: Finding Sparse, Trainable Neural
           Networks},
  author = {Jonathan Frankle and Michael Carbin},
  year = {2019},
  eprint = {1803.03635},
  archivePrefix = {arXiv},
  primaryClass = {cs.LG},
  url = {https://arxiv.org/abs/1803.03635},
}

@article{liang21a,
  title = {Pruning and quantization for deep neural network acceleration: A
           survey},
  journal = {Neurocomputing},
  volume = {461},
  pages = {370-403},
  year = {2021},
  issn = {0925-2312},
  doi = {https://doi.org/10.1016/j.neucom.2021.07.045},
  url = {https://www.sciencedirect.com/science/article/pii/S0925231221010894},
  author = {Tailin Liang and John Glossner and Lei Wang and Shaobo Shi and
            Xiaotong Zhang},
}

@inproceedings{blalock20a,
  author = {Blalock, Davis and Gonzalez Ortiz, Jose Javier and Frankle, Jonathan
            and Guttag, John},
  booktitle = {Proceedings of Machine Learning and Systems},
  editor = {I. Dhillon and D. Papailiopoulos and V. Sze},
  pages = {129--146},
  title = {What is the State of Neural Network Pruning?},
  url = {
         https://proceedings.mlsys.org/paper_files/paper/2020/file/6c44dc73014d66ba49b28d483a8f8b0d-Paper.pdf
         },
  volume = {2},
  year = {2020},
}

@misc{behrends22a,
  title = {Method and systems for identifying a sequence of monomer units of a
           biological or synthetic heteropolymer},
  author = {Behrends, Jan C. and Ensslen, Tobias},
  year = {Patent US20240077491A1, March 7, 2024},
}

@article{bakshloo22a,
  title = {Nanopore-Based Protein Identification},
  volume = {144},
  issn = {0002-7863, 1520-5126},
  url = {https://pubs.acs.org/doi/10.1021/jacs.1c11758},
  doi = {10.1021/jacs.1c11758},
  pages = {2716--2725},
  number = {6},
  journal = {Journal of the American Chemical Society},
  shortjournal = {J. Am. Chem. Soc.},
  author = {Afshar Bakshloo, Mazdak and Kasianowicz, John J. and
            Pastoriza-Gallego, Manuela and Mathé, Jérôme and Daniel, Régis and
            Piguet, Fabien and Oukhaled, Abdelghani},
  year = {2022},
}

@article{muradeli20a,
  title = {ssqueezepy},
  author = {John Muradeli},
  journal = {GitHub. Note: https://github.com/OverLordGoldDragon/ssqueezepy/},
  year = {2020},
  doi = {10.5281/zenodo.5080508},
}

@article{zhang24a,
  author = {Zhang, Ming and Tang, Chao and Wang, Zichun and Chen, Shanchuan and
            Zhang, Dan and Li, Kaiju and Sun, Ke and Zhao, Changjian and Wang, Yu
            and Xu, Mengying and Dai, Lunzhi and Lu, Guangwen and Shi, Hubing and
            Ren, Haiyan and Chen, Lu and Geng, Jia},
  da = {2024/04/01},
  doi = {10.1038/s41592-024-02208-7},
  id = {Zhang2024},
  isbn = {1548-7105},
  journal = {Nature Methods},
  number = {4},
  pages = {609--618},
  title = {Real-time detection of 20 amino acids and discrimination of
           pathologically relevant peptides with functionalized nanopore},
  ty = {JOUR},
  url = {https://doi.org/10.1038/s41592-024-02208-7},
  volume = {21},
  year = {2024},
}

@article{wang23b,
  author = {Wang, Fushi and Zhao, Chunxiao and Zhao, Pinlong and Chen, Fanfan
            and Qiao, Dan and Feng, Jiandong},
  da = {2023/05/20},
  doi = {10.1038/s41467-023-38627-x},
  id = {Wang2023},
  isbn = {2041-1723},
  journal = {Nature Communications},
  number = {1},
  pages = {2895},
  title = {MoS2 nanopore identifies single amino acids with sub-1 Dalton
           resolution},
  ty = {JOUR},
  url = {https://doi.org/10.1038/s41467-023-38627-x},
  volume = {14},
  year = {2023},
}

@article{boukhet17a,
  title = {Translocation of {{Precision Polymers}} through {{Biological
           Nanopores}}},
  author = {Boukhet, Mordjane and K{\"o}nig, Niklas Felix and Ouahabi, Abdelaziz
            Al and Baaken, Gerhard and Lutz, Jean-Fran{\c c}ois and Behrends, Jan
            C.},
  year = 2017,
  month = dec,
  journal = {Macromolecular Rapid Communications},
  volume = {38},
  number = {24},
  pages = {1700680},
  issn = {1022-1336, 1521-3927},
  doi = {10.1002/marc.201700680},
  urldate = {2026-01-22},
  copyright = {http://onlinelibrary.wiley.com/termsAndConditions\#vor},
  langid = {english},
}

@phdthesis{ensslen24a,
  title = {Mechanistic Principles of High- Resolution Discrimination of Polymers
           with Membrane Proteins},
  author = {Ensslen, Tobias},
  year = 2024,
  address = {Freiburg},
  langid = {english},
  school = {Albert-Ludwigs-Universit\"at Freiburg},
  note = {\url{https://www.doi.org/10.6094/UNIFR/269188}}
}

@phdthesis{boukhet18a,
  title = {Discrimination and {{Sequencing}} of {{Polymers}} with {{Biological
           Nanopores}}},
  author = {Boukhet, Mordjane},
  year = 2018,
  month = nov,
  number = {2018CERG0984},
  urldate = {2026-01-22},
  address = {Cergy-Pontoise},
  school = {Universit\'e de Cergy Pontoise ; Albert-Ludwigs-Universit\"at
            Freiburg},
  note = {\url{https://theses.hal.science/tel-02311628v1}}
}

@article{hossbach25a,
  title = {Peptide Classification from Statistical Analysis of Nanopore Sensing
           Experiments},
  author = {Ho{\ss}bach, Julian and Tovey, Samuel and Ensslen, Tobias and
            Behrends, Jan C. and Holm, Christian},
  year = {2025},
  month = feb,
  journal = {The Journal of Chemical Physics},
  volume = {162},
  number = {8},
  pages = {084107},
  issn = {0021-9606},
  doi = {10.1063/5.0250399},
  urldate = {2025-02-25},
}

@article{ouldali20a,
  title = {Electrical Recognition of the Twenty Proteinogenic Amino Acids Using
           an Aerolysin Nanopore},
  author = {Ouldali, Hadjer and Sarthak, Kumar and Ensslen, Tobias and Piguet,
            Fabien and Manivet, Philippe and Pelta, Juan and Behrends, Jan C. and
            Aksimentiev, Aleksei and Oukhaled, Abdelghani},
  year = {2020-02, 2020},
  journal = {Nature Biotechnology},
  volume = {38},
  number = {2},
  pages = {176--181},
  issn = {1546-1696},
  doi = {10.1038/s41587-019-0345-2},
  urldate = {2023-01-11},
  copyright = {2019 The Author(s), under exclusive licence to Springer Nature
               America, Inc.},
  langid = {english},
}

@article{ritmejeris24a,
  title = {Single-Molecule Protein Sequencing with Nanopores},
  author = {Ritmejeris, Justas and Chen, Xiuqi and Dekker, Cees},
  year = {2024},
  month = nov,
  journal = {Nature Reviews Bioengineering},
  volume = {3},
  number = {4},
  pages = {303--316},
  issn = {2731-6092},
  doi = {10.1038/s44222-024-00260-8},
  urldate = {2025-08-02},
  langid = {english},
}

@article{brinkerhoff21a,
  title = {Multiple Rereads of Single Proteins at Single--Amino Acid Resolution
           Using Nanopores},
  author = {Brinkerhoff, Henry and Kang, Albert S. W. and Liu, Jingqian and
            Aksimentiev, Aleksei and Dekker, Cees},
  year = {2021},
  journal = {Science},
  volume = {374},
  number = {6574},
  pages = {1509--1513},
  doi = {10.1126/science.abl4381},
}

@inproceedings{lundberg17a,
  title = {A Unified Approach to Interpreting Model Predictions},
  booktitle = {Proceedings of the 31st {{International Conference}} on {{Neural
               Information Processing Systems}}},
  author = {Lundberg, Scott M and Lee, Su-In},
  editor = {Guyon, I. and Luxburg, U. Von and Bengio, S. and Wallach, H. and
            Fergus, R. and Vishwanathan, S. and Garnett, R.},
  year = {2017},
  volume = {30},
  publisher = {Curran Associates, Inc.},
  location = {Long Beach, California, USA},
  url = {
         https://proceedings.neurips.cc/paper_files/paper/2017/file/8a20a8621978632d76c43dfd28b67767-Paper.pdf
         },
}

@misc{kokhlikyan20a-pre,
  title = {Captum: A unified and generic model interpretability library for
           PyTorch},
  author = {Narine Kokhlikyan and Vivek Miglani and Miguel Martin and Edward
            Wang and Bilal Alsallakh and Jonathan Reynolds and Alexander Melnikov
            and Natalia Kliushkina and Carlos Araya and Siqi Yan and Orion
            Reblitz-Richardson},
  year = {2020},
  eprint = {2009.07896},
  archivePrefix = {arXiv},
  primaryClass = {cs.LG},
  url = {https://arxiv.org/abs/2009.07896},
}

@misc{he15a,
  title = {Deep Residual Learning for Image Recognition},
  author = {Kaiming He and Xiangyu Zhang and Shaoqing Ren and Jian Sun},
  year = {2015},
  eprint = {1512.03385},
  archivePrefix = {arXiv},
  primaryClass = {cs.CV},
  url = {https://arxiv.org/abs/1512.03385},
}

@inproceedings{sutskever13a,
  title = {On the Importance of Initialization and Momentum in Deep Learning},
  booktitle = {Proceedings of the 30th International Conference on Machine
               Learning},
  author = {Sutskever, Ilya and Martens, James and Dahl, George and Hinton,
            Geoffrey},
  editor = {Dasgupta, Sanjoy and McAllester, David},
  date = {2013-06-17/2013-06-19},
  year = {2013},
  series = {Proceedings of Machine Learning Research},
  volume = {28},
  pages = {1139--1147},
  publisher = {PMLR},
  location = {Atlanta, Georgia, USA},
  url = {https://proceedings.mlr.press/v28/sutskever13.html},
}

@misc{xie17a,
  title = {Aggregated Residual Transformations for Deep Neural Networks},
  author = {Saining Xie and Ross Girshick and Piotr Dollár and Zhuowen Tu and
            Kaiming He},
  year = {2017},
  eprint = {1611.05431},
  archivePrefix = {arXiv},
  primaryClass = {cs.CV},
  url = {https://arxiv.org/abs/1611.05431},
}

@inproceedings{hossain99a,
  author = {Hossain, M.I. and Liu, J. and Lee, R.},
  booktitle = {IEEE SMC'99 Conference Proceedings. 1999 IEEE International
               Conference on Systems, Man, and Cybernetics (Cat. No.99CH37028)},
  title = {A study of multilingual speech features: perceptive scalogram based
           on wavelet analysis},
  year = {1999},
  volume = {2},
  number = {},
  pages = {178-183 vol.2},
  doi = {10.1109/ICSMC.1999.825229},
}

@article{sejdic08a,
  author = {Sejdic, Ervin and Djurovic, Igor and Stankovic, LJubisa},
  journal = {IEEE Transactions on Signal Processing},
  title = {Quantitative Performance Analysis of Scalogram as Instantaneous
           Frequency Estimator},
  year = {2008},
  volume = {56},
  number = {8},
  pages = {3837-3845},
  doi = {10.1109/TSP.2008.924856},
}

@article{cao24a,
  title = {Deep {{Learning-Assisted Single-Molecule Detection}} of {{Protein
           Post-translational Modifications}} with a {{Biological Nanopore}}},
  author = {Cao, Chan and Magalhães, Pedro and Krapp, Lucien F. and Bada Juarez,
            Juan F. and Mayer, Simon Finn and Rukes, Verena and Chiki, Anass and
            Lashuel, Hilal A. and Dal Peraro, Matteo},
  date = {2024-01-16},
  year = {2024},
  journal = {ACS Nano},
  journaltitle = {ACS Nano},
  shortjournal = {ACS Nano},
  volume = {18},
  number = {2},
  pages = {1504--1515},
  publisher = {American Chemical Society},
  issn = {1936-0851},
  doi = {10.1021/acsnano.3c08623},
}

@incollection{piguet21a,
  title = {Chapter {{Twenty}} - {{Pore-forming}} Toxins as Tools for Polymer
           Analytics: {{From}} Sizing to Sequencing},
  booktitle = {Pore-{{Forming Toxins}}},
  author = {Piguet, Fabien and Ensslen, Tobias and Bakshloo, Mazdak A. and
            Talarimoghari, Monasadat and Ouldali, Hadjer and Baaken, Gerhard and
            Zaitseva, Ekaterina and Pastoriza-Gallego, Manuela and Behrends, Jan
            C. and Oukhaled, Abdelghani},
  editor = {Heuck, Alejandro P.},
  year = {2021},
  series = {Methods in {{Enzymology}}},
  volume = {649},
  pages = {587--634},
  publisher = {Academic Press},
  issn = {0076-6879},
  doi = {10.1016/bs.mie.2021.01.017},
}

@article{garcia06a,
  author = {García, Thorsten Tepper},
  title = "{Voigt profile fitting to quasar absorption lines: an analytic
           approximation to the Voigt–Hjerting function}",
  journal = {Monthly Notices of the Royal Astronomical Society},
  volume = {369},
  number = {4},
  pages = {2025-2035},
  year = {2006},
  month = {07},
  issn = {0035-8711},
  doi = {10.1111/j.1365-2966.2006.10450.x},
  url = {https://doi.org/10.1111/j.1365-2966.2006.10450.x},
  eprint = {
            https://academic.oup.com/mnras/article-pdf/369/4/2025/3829818/mnras0369-2025.pdf
            },
}

@misc{paszke19a,
  title = {PyTorch: An Imperative Style, High-Performance Deep Learning Library},
  author = {Adam Paszke and Sam Gross and Francisco Massa and Adam Lerer and
            James Bradbury and Gregory Chanan and Trevor Killeen and Zeming Lin
            and Natalia Gimelshein and Luca Antiga and Alban Desmaison and
            Andreas Köpf and Edward Yang and Zach DeVito and Martin Raison and
            Alykhan Tejani and Sasank Chilamkurthy and Benoit Steiner and Lu Fang
            and Junjie Bai and Soumith Chintala},
  year = {2019},
  eprint = {1912.01703},
  archivePrefix = {arXiv},
  primaryClass = {cs.LG},
  url = {https://arxiv.org/abs/1912.01703},
}

@misc{falcon19a,
  author = {Falcon, William and {The PyTorch Lightning team}},
  doi = {10.5281/zenodo.3828935},
  license = {Apache-2.0},
  month = mar,
  title = {{PyTorch Lightning}},
  url = {https://github.com/Lightning-AI/lightning},
  version = {1.4},
  year = {2019},
}

@inproceedings{deng09a,
  author = {Deng, Jia and Dong, Wei and Socher, Richard and Li, Li-Jia and Kai
            Li and Li Fei-Fei},
  booktitle = {2009 IEEE Conference on Computer Vision and Pattern Recognition},
  title = {ImageNet: A large-scale hierarchical image database},
  year = {2009},
  volume = {},
  number = {},
  pages = {248-255},
  doi = {10.1109/CVPR.2009.5206848},
}

@inproceedings{hendrycks19a,
  title = {Using Pre-Training Can Improve Model Robustness and Uncertainty},
  author = {Hendrycks, Dan and Lee, Kimin and Mazeika, Mantas},
  booktitle = {Proceedings of the 36th International Conference on Machine
               Learning},
  pages = {2712--2721},
  year = {2019},
  editor = {Chaudhuri, Kamalika and Salakhutdinov, Ruslan},
  volume = {97},
  series = {Proceedings of Machine Learning Research},
  month = {09--15 Jun},
  publisher = {PMLR},
  url = {https://proceedings.mlr.press/v97/hendrycks19a.html},
}

@misc{zhong17a,
  title = {Random Erasing Data Augmentation},
  author = {Zhun Zhong and Liang Zheng and Guoliang Kang and Shaozi Li and Yi
            Yang},
  year = {2017},
  eprint = {1708.04896},
  archivePrefix = {arXiv},
  primaryClass = {cs.CV},
  url = {https://arxiv.org/abs/1708.04896},
}

@misc{izmailov19a,
  title = {Averaging Weights Leads to Wider Optima and Better Generalization},
  author = {Pavel Izmailov and Dmitrii Podoprikhin and Timur Garipov and Dmitry
            Vetrov and Andrew Gordon Wilson},
  year = {2019},
  eprint = {1803.05407},
  archivePrefix = {arXiv},
  primaryClass = {cs.LG},
  url = {https://arxiv.org/abs/1803.05407},
}

@misc{athiwaratkun19a,
  title = {There Are Many Consistent Explanations of Unlabeled Data: Why You
           Should Average},
  author = {Ben Athiwaratkun and Marc Finzi and Pavel Izmailov and Andrew Gordon
            Wilson},
  year = {2019},
  eprint = {1806.05594},
  archivePrefix = {arXiv},
  primaryClass = {cs.LG},
  url = {https://arxiv.org/abs/1806.05594},
}

@misc{he21a,
  title = {Masked Autoencoders Are Scalable Vision Learners},
  author = {Kaiming He and Xinlei Chen and Saining Xie and Yanghao Li and Piotr
            Dollár and Ross Girshick},
  year = {2021},
  eprint = {2111.06377},
  archivePrefix = {arXiv},
  primaryClass = {cs.CV},
  url = {https://arxiv.org/abs/2111.06377},
}

@inproceedings{erhan10a,
  title = {Why Does Unsupervised Pre-training Help Deep Learning?},
  author = {Erhan, Dumitru and Courville, Aaron and Bengio, Yoshua and Vincent,
            Pascal},
  booktitle = {Proceedings of the Thirteenth International Conference on
               Artificial Intelligence and Statistics},
  pages = {201--208},
  year = {2010},
  editor = {Teh, Yee Whye and Titterington, Mike},
  volume = {9},
  series = {Proceedings of Machine Learning Research},
  address = {Chia Laguna Resort, Sardinia, Italy},
  month = {13--15 May},
  publisher = {PMLR},
  url = {https://proceedings.mlr.press/v9/erhan10a.html},
}

@misc{devlin19a,
  title = {BERT: Pre-training of Deep Bidirectional Transformers for Language
           Understanding},
  author = {Jacob Devlin and Ming-Wei Chang and Kenton Lee and Kristina
            Toutanova},
  year = {2019},
  eprint = {1810.04805},
  archivePrefix = {arXiv},
  primaryClass = {cs.CL},
  url = {https://arxiv.org/abs/1810.04805},
}

@misc{loshchilov17a,
  title = {SGDR: Stochastic Gradient Descent with Warm Restarts},
  author = {Ilya Loshchilov and Frank Hutter},
  year = {2017},
  eprint = {1608.03983},
  archivePrefix = {arXiv},
  primaryClass = {cs.LG},
  url = {https://arxiv.org/abs/1608.03983},
}

@misc{hendrycks23a,
  title = {Gaussian Error Linear Units (GELUs)},
  author = {Dan Hendrycks and Kevin Gimpel},
  year = {2023},
  eprint = {1606.08415},
  archivePrefix = {arXiv},
  primaryClass = {cs.LG},
  url = {https://arxiv.org/abs/1606.08415},
}

@inproceedings{nagel22a,
  title = {Overcoming Oscillations in Quantization-Aware Training},
  author = {Nagel, Markus and Fournarakis, Marios and Bondarenko, Yelysei and
            Blankevoort, Tijmen},
  booktitle = {Proceedings of the 39th International Conference on Machine
               Learning},
  pages = {16318--16330},
  year = {2022},
  editor = {Chaudhuri, Kamalika and Jegelka, Stefanie and Song, Le and
            Szepesvari, Csaba and Niu, Gang and Sabato, Sivan},
  volume = {162},
  series = {Proceedings of Machine Learning Research},
  month = {17--23 Jul},
  publisher = {PMLR},
  url = {https://proceedings.mlr.press/v162/nagel22a.html},
}

@misc{yu22a,
  title = {CoCa: Contrastive Captioners are Image-Text Foundation Models},
  author = {Jiahui Yu and Zirui Wang and Vijay Vasudevan and Legg Yeung and
            Mojtaba Seyedhosseini and Yonghui Wu},
  year = {2022},
  eprint = {2205.01917},
  archivePrefix = {arXiv},
  primaryClass = {cs.CV},
  url = {https://arxiv.org/abs/2205.01917},
}

@misc{ridnik21a,
  title = {ImageNet-21K Pretraining for the Masses},
  author = {Tal Ridnik and Emanuel Ben-Baruch and Asaf Noy and Lihi Zelnik-Manor
            },
  year = {2021},
  eprint = {2104.10972},
  archivePrefix = {arXiv},
  primaryClass = {cs.CV},
  url = {https://arxiv.org/abs/2104.10972},
}

@article{noakes19a,
  author = {Noakes, Matthew T. and Brinkerhoff, Henry and Laszlo, Andrew H. and
            Derrington, Ian M. and Langford, Kyle W. and Mount, Jonathan W. and
            Bowman, Jasmine L. and Baker, Katherine S. and Doering, Kenji M. and
            Tickman, Benjamin I. and Gundlach, Jens H.},
  date = {2019/06/01},
  doi = {10.1038/s41587-019-0096-0},
  id = {Noakes2019},
  isbn = {1546-1696},
  journal = {Nature Biotechnology},
  number = {6},
  pages = {651--656},
  title = {Increasing the accuracy of nanopore DNA sequencing using a
           time-varying cross membrane voltage},
  url = {https://doi.org/10.1038/s41587-019-0096-0},
  volume = {37},
  year = {2019},
}

@article{he21b,
  author = {He, Yuhui and Tsutsui, Makusu and Zhou, Yue and Miao, Xiang-Shui},
  date = {2021/06/11},
  doi = {10.1038/s41427-021-00313-z},
  id = {He2021},
  isbn = {1884-4057},
  journal = {NPG Asia Materials},
  number = {1},
  pages = {48},
  title = {Solid-state nanopore systems: from materials to applications},
  url = {https://doi.org/10.1038/s41427-021-00313-z},
  volume = {13},
  year = {2021},
}

@article{ratinho25a,
  author = {Ratinho, Laura and Meyer, Nathan and Greive, Sandra and Cressiot,
            Benjamin and Pelta, Juan},
  date = {2025/04/04},
  doi = {10.1038/s41467-025-58509-8},
  id = {Ratinho2025},
  isbn = {2041-1723},
  journal = {Nature Communications},
  number = {1},
  pages = {3211},
  title = {Nanopore sensing of protein and peptide conformation for
           point-of-care applications},
  url = {https://doi.org/10.1038/s41467-025-58509-8},
  volume = {16},
  year = {2025},
}
\end{document}